\documentclass{article}

\usepackage[utf8]{inputenc}
\usepackage[T1]{fontenc}
\usepackage{PRIMEarxiv}
\usepackage{authblk}
\usepackage{amsmath, amssymb, amsfonts, bm}
\usepackage{amsthm}
\usepackage{enumitem} 
\usepackage{graphicx}
\usepackage{subcaption}
\usepackage{booktabs}
\usepackage{multirow}
\usepackage{longtable}

\usepackage{amsmath, amssymb, amsthm}
 
\usepackage{tikz}
\usetikzlibrary{positioning, calc, fit}

\usepackage{tabularx}
\usepackage{adjustbox}
\usepackage{siunitx}
\usepackage{algorithm}
\usepackage{hyperref}
\usepackage{algorithmicx}
\usepackage{algpseudocode}
\usepackage{mathrsfs}
\usepackage{xcolor}
\usepackage{soul}
\sethlcolor{yellow}
\usepackage{textcomp}
\usepackage{comment}
\usepackage{url}
\usepackage{natbib}
\soulregister\cite7
\soulregister\citep7
\soulregister\citet7
\soulregister\ref7
\soulregister\pageref7
\usepackage{fancyhdr}
\usepackage{microtype}
\usepackage[title]{appendix}
\usepackage{setspace}
\usepackage[left]{lineno}
\usepackage{hyperref}
\usepackage[normalem]{ulem}
\usepackage{adjustbox}  
\usepackage{pgfplots}
\usepgfplotslibrary{groupplots}

\usepackage{comment}
\usepackage{float}
\usepackage{wrapfig}
\usepackage{caption}
\usepackage{tikz}
\usetikzlibrary{positioning, arrows.meta, calc, fit, backgrounds}
\usepackage{pgfplots}
\usepgfplotslibrary{groupplots}
\pgfplotsset{compat=1.18}

\usepackage{xurl}             
\usepackage{hyperref}         
\hypersetup{
  breaklinks=true,            
  colorlinks=true,            
  urlcolor=blue               
}

\usepackage{rotating}

\hypersetup{
  colorlinks=true,                      
  citecolor=blue,                       
  linkcolor=blue,                       
  urlcolor=blue                         
}

\microtypecontext{spacing=nonfrench}

\title{Agentic Physical AI toward a Domain-Specific Foundation Model for Energy Systems: A Case Study on Nuclear Reactor Control}

\author[1,2]{Yoon Pyo Lee}
\author[1]{Samrendra Roy}
\author[1]{Kazuma Kobayashi}
\author[6]{Sajedul Talukder}
\author[3, 7]{Diab Abueidda}
\author[3]{Seid Koric}
\author[4,5]{Souvik Chakraborty}
\author[1,3]{Syed Bahauddin Alam\textsuperscript{*}}

\affil[1]{The Grainger College of Engineering, Nuclear, Plasma \& Radiological Engineering, University of Illinois Urbana-Champaign, Urbana, IL, USA
}
\affil[2]{Department of Nuclear Engineering, Hanyang University, Republic of Korea}

\affil[3]{National Center for Supercomputing Applications, Urbana, IL, USA
}

\affil[4]{Department of Applied Mechanics, Indian Institute of Technology Delhi, New Delhi, India
}
\affil[5]{Yardi School of Artificial Intelligence, Indian Institute of Technology Delhi}
\affil[6]{University of Texas - El Paso, University Ave, El Paso, USA}
\affil[7]{Civil and Urban Engineering Department, New York University Abu Dhabi, UAE}

\affil[*]{Corresponding author: \href{mailto:alams@illinois.edu}{alams@illinois.edu}}

\begin{document}
\setcounter{page}{1}
\maketitle
\pagestyle{fancy}


\vspace{-3mm}

\begin{abstract}
The prevailing paradigm in AI for physical systems: scaling general-purpose foundation models toward universal multimodal reasoning, confronts a barrier at the control interface. Frontier vision--language models achieve only 50--53\% accuracy on basic quantitative physics tasks, behaving as approximate guessers that preserve semantic plausibility while violating physical constraints. Safety-critical control demands outcome-space guarantees over executed actions, not parameter-space imitation. Here we present a pathway \emph{toward} domain-specific foundation models through compact language models operating as Agentic Physical AI: policy optimization driven by physics-based simulator validation rather than perceptual inference. We train a 360M-parameter model on synthetic nuclear reactor scenarios scaled from $10^3$ to $10^5$ examples. Scaling produces strong, regime-dependent reliability gains under nominal simulated conditions, with variance collapse of approximately $500\times$ and elimination of $>10\%$ terminal-power excursions on the sampled distribution. Despite balanced exposure to four actuation families, the model concentrates 95\% of runtime execution on a single-bank strategy, without reinforcement learning or reward engineering. Representations transfer across simulators without architectural change. We position the system as a candidate decision component within a verification, monitoring, and defense-in-depth architecture, not as a stand-alone safety solution: the demonstrated behavior speaks to closed-loop reliability on a single-step task in simulation and does not yet address off-nominal operation, sensor faults, or uncertainty quantification.
\end{abstract}


\section*{Introduction}

Nuclear energy stands at an inflection point. Decarbonization targets demand flexible, load-following support to renewable-heavy grids, while small modular reactors (SMRs) promise deployment in remote environments with reduced staffing, supporting mission profiles (military bases, disaster relief, deep-space exploration) that legacy plants were never designed for. Digital twins and high-fidelity simulators now generate synthetic operational data at unprecedented scale~\cite{imron2019development,huff2015pyrk,kobayashi2024deeponet,hossain2025virtual}. Yet nuclear control architectures remain frozen in a 1970s paradigm: plant-specific proportional-integral-derivative (PID) loops, manually tuned supervisory logic, and model-predictive controllers that require online optimization and cannot transfer across reactor designs~\cite{hu2022model,huang2023model,mostafa2024improved}.
The nuclear industry has access to orders of magnitude more synthetic training data than it can exploit, yet control design remains a bespoke, plant-by-plant exercise—a paradigm mismatch rather than a computational bottleneck. Machine learning has delivered transformative gains in perception~\cite{lee2021convolutional}, diagnostics~\cite{chae2022graph,kobayashi2024explainable,hossain2024sensor}, 
and document processing~\cite{al2024scalable,lee2025large}, but has been systematically excluded from \emph{actuation}: the task that matters most. The reason is fundamental: conventional supervised learning optimizes Euclidean distance in parameter space, implicitly assuming safety is a smooth, convex function of control inputs. It is not. Reactor safety resides on a low-dimensional, disjoint solution manifold embedded within a high-dimensional, brittle action space. Minimizing parameter error does not guarantee convergence to this manifold; outcome-centric scaling, by contrast, directly rewards strategies that succeed in closed-loop execution and therefore concentrates probability mass on the reliable sub-region of the action space.

The urgency of solving this paradigm mismatch has been elevated to a national imperative. The Genesis Mission: A National Mission to Accelerate Science Through Artificial Intelligence (Executive Order 14363) explicitly mandates ``domain-specific foundation models'' (Sec.~3(a)(iv)) and ``AI agents to explore design spaces'' (Sec.~3(a)(ii)) to secure energy dominance~\cite{whitehouse2025genesisEO,whitehouse2025genesisFS}. The concurrent National Academies of Sciences, Engineering, and Medicine (NASEM) study ``Foundation Models for Scientific Discovery and Innovation: Opportunities Across the Department of Energy and the Scientific Enterpris''~\cite{NAP29212} identifies the alignment gap between general-purpose AI capabilities and the ``mission-critical constraints'' of United States Department of Energy (DOE) infrastructure as the central challenge. Realizing the power of foundation models in safety-critical domains requires ``alignment on Verification, Validation, and Uncertainty Quantification (VVUQ)'', a standard that current ``plausible-sounding'' generative models fail to meet. Nuclear reactor control, as the most safety-critical and constraint-rich domain within the DOE enterprise, tests whether data-scaled foundation models can meet VVUQ requirements while delivering the operational flexibility that SMR deployment and grid integration demand.

Three structural barriers prevent learning-enabled actuation in nuclear systems and block the nuclear sector from fulfilling the Genesis Mission's directive for agentic AI. These barriers are fundamental properties of how control is currently conceived, trained, and validated, creating conflicts between legacy control paradigms and the requirements for domain-specific foundation models:

\begin{enumerate}[itemsep=0pt]
\item The Training Objective Mismatch (Many-to-One Actuation Spaces):
In pressurized water reactors, multiple rod actuation sequences (single-bank maneuvers, simultaneous two-bank insertions, sequential patterns) can all achieve identical power transitions. Conventional supervised learning optimizes parameter-level losses (mean absolute error on rod positions), implicitly assuming one correct answer per state. But when multiple actuation strategies yield physically equivalent outcomes, minimizing parameter error is the wrong objective. A model that strictly imitates a single reference action may perform worse than one that learns to select among several admissible strategies based on reliability or equipment constraints. Current training protocols penalize this flexibility as variance rather than recognizing it as the structure of the domain. A foundation-model approach for the DOE enterprise must therefore learn the \emph{topology of the admissible solution space}, not merely mimic a single operator's choice.

\item The Exploration Risk Barrier (Offline Learning Under Safety Constraints):
Reinforcement learning has been investigated for nuclear load-following and microreactor control, often improving transient tracking relative to PID baselines in simplified environments~\cite{gong2024possibilities,tunkle2025nuclear,radaideh2026multistep,degrave2022magnetic}. However, these approaches require large volumes of environment interaction, are sensitive to reward shaping, and face nontrivial challenges in transferring policies across reactor designs. Even when embedded within model-based or constrained frameworks~\cite{hu2022model,xiao2022neural,yin2023design}, their deployment pathway remains unclear for licensing-grade actuation. Current RL paradigms are fundamentally incompatible with nuclear safety culture: the industry cannot deploy agents that require online learning to become safe, yet cannot simulate every possible edge case offline. The solution lies in the NASEM-recommended strategy of leveraging ``massive data sets'' to train offline agents that internalize safety constraints \emph{before} deployment, eliminating the need for hazardous online exploration.

\item The Plausibility-Physics Gap (From Verbal Inference to Closed-Loop Execution):
Recent benchmarks show that state-of-the-art vision-language models rely on memorized priors rather than faithfully using provided numeric and visual evidence, leading to brittle and input-unfaithful estimates even in controlled laboratory tasks~\cite{puyin2025quantiphy,physbench2025,star2024,contphy2024}. These models are evaluated in open-loop question-answer settings and rewarded for plausible numbers rather than physically verified consequences. In nuclear control, a model that sounds right numerically but fails when executed in a reactor simulator is worse than useless. The Genesis Mission mandates ``AI-enabled predictive models'' rigorous enough for national security sites. A model that produces numerically plausible answers yet cannot synthesize safe control trajectories under actuator constraints is insufficient for safety-critical deployment.
\end{enumerate}

These barriers are structural properties of the dominant learning paradigms. The nuclear sector possesses both data abundance (synthetic corpora from digital twins~\cite{imron2019development,huff2015pyrk,kobayashi2024deeponet}) and institutional mandate (Genesis Mission Sec.~3(a)(v): ``federally curated and open scientific datasets''), yet control design remains handcrafted at the loop level while evidence about safe actuation now exists at the corpus level.

Agentic Physical AI provides a framework in which compact language models (360M parameters) learn control policies through outcome-centric validation in physics-based simulators rather than parameter-space imitation. Scaling synthetic data from 1,000 to 100,000 scenarios induces strong scaling behavior with regime-dependent gains: success rates at $\pm$1\% tolerance jump from 6.0\% to 92.0\%, variance collapses 500-fold, and the model autonomously concentrates 95\% of execution on physically robust single-bank strategies despite balanced training exposure to four actuation families. This emergent policy distillation arises without reinforcement learning or reward engineering, driven solely by outcome-level success under physical execution.

The framework learns the topology of safe actuation offline, proposes diverse strategies without exploration risk, and filters candidates through closed-loop physics execution rather than numerical plausibility. These structural barriers within nuclear control are compounded by fundamental limitations in the general-purpose AI architectures currently dominating the field.

Learning-enabled approaches to nuclear reactor control have been explored since at least the 1990s, with neural-network control methods for nuclear power plants surveyed by Zhou and Tan~\cite{zhou2023review}. Representative early and mid-stage examples include recurrent-neural-network and fuzzy-system core controllers for load-following operation~\cite{boroushaki2003intelligent}, fuzzy model-predictive power control for pressurized water reactors~\cite{na2006design}, and model-predictive load-following control with discrete optimization of control-rod speed~\cite{kim2014design}. Reinforcement learning has more recently been investigated for load-following and microreactor control~\cite{gong2024possibilities,nguyen2024reinforcement,tunkle2025nuclear,radaideh2026multistep}, with several studies reporting improved transient tracking relative to fixed-gain PID baselines. Model-predictive control combined with neural surrogates~\cite{hu2022model,xiao2022neural,yin2023design} and Koopman- or SINDy-based reduced-order identification~\cite{kaiser2018sparse} have been used to enable fast constrained optimization. Within the broader operational stack, supervised learning has been applied to fault detection, signal validation, and condition monitoring~\cite{lee2021convolutional,chae2022graph,kobayashi2024explainable,hossain2024sensor}, and large language models have been used for regulatory text processing and operator decision support~\cite{al2024scalable,lee2025large}. Two gaps in this literature motivate the present work: (i) most learning-enabled controllers are trained and reported per plant, with limited evidence of cross-reactor transfer; and (ii) the dominant evaluation paradigm remains parameter-space loss or single-trajectory tracking error rather than population-level, outcome-centric closed-loop reliability across many independent scenarios. We target these two gaps directly: a single backbone is adapted to two distinct reactor-physics simulators (KOMODO spatial neutronics; PyRK point kinetics) via a shared two-phase curriculum, and evaluation is performed across thousands of independent closed-loop runs rather than on a small validation trajectory.

Gap 1: From passive quantitative perception to closed-loop physical control.
Recent benchmarks in quantitative physical reasoning reveal a fundamental limitation of vision--language and multimodal foundation models: even when provided pixel-accurate video and explicit physical priors, state-of-the-art systems achieve only 50--53\% accuracy on basic kinematic inference tasks, exhibiting behavior consistent with approximate guessing rather than measurement~\cite{puyin2025quantiphy,physbench2025}. More critically, these evaluations treat physical reasoning as a passive inference problem, predicting a scalar quantity from a short observation, without considering how such inferences would be used to generate actions, maintain stability, or respect safety envelopes~\cite{star2024,contphy2024}. In contrast, nuclear reactor operation is inherently a closed-loop control problem, where the correctness of a model is determined not by the plausibility of an estimate, but by the physical consequences of the actions it proposes over time. A model that produces numerically plausible answers yet cannot synthesize safe control trajectories under actuator constraints is insufficient for safety-critical deployment.

Gap 2: Lack of input-faithful, physics-constrained reasoning.
A deeper diagnosis emerging from recent physical reasoning benchmarks is the lack of input faithfulness in current foundation models. When visual evidence is removed and only textual priors are provided, performance degrades only marginally; when those priors are counterfactually rescaled, predictions often fail catastrophically, reverting to typical world statistics rather than honoring the modified inputs~\cite{puyin2025quantiphy}. Similar brittleness has been observed across multimodal reasoning benchmarks, where semantic coherence is preserved at the expense of numerical and physical consistency~\cite{videophy2025,deepphy2025}. Such models rely predominantly on internalized parametric priors rather than enforcing physical constraints derived from sensor data—tolerable for descriptive tasks but fundamentally incompatible with nuclear reactor control, where calibrated sensor streams, updated operating conditions, and revised safety margins must act as hard constraints on decision-making.

Gap 3: From low-dimensional kinematics to high-dimensional safety manifolds.
Existing physical reasoning benchmarks deliberately restrict scope to low-dimensional kinematic quantities:object size, velocity, or acceleration, in short, visually simple scenes with rigid bodies, translational motion, and fixed cameras~\cite{puyin2025quantiphy,clevrer2020}. Nuclear reactor control, by contrast, operates in a high-dimensional, tightly coupled state space governed by neutron kinetics, thermal--hydraulic feedback, delayed neutron dynamics, and actuator rate limits. Safety is not determined by the accuracy of any single inferred quantity, but by whether system trajectories remain within a narrow feasible manifold over time. Consequently, success in nuclear control requires models that internalize the geometry of this manifold and synthesize actions that keep trajectories within it, rather than models optimized to regress isolated physical quantities. This distinction motivates a shift from perception-centric physical reasoning toward agentic, outcome-validated control in which physics execution: not numerical plausibility, defines correctness, necessitating domain-specific foundation models rather than general-purpose multimodal systems.

The central question of this work is not whether AI can assist nuclear control (that has been demonstrated in narrowly scoped perception and diagnostics tasks). The question is whether AI can \emph{replace the current paradigm of plant-specific, optimization-based control design} with a reusable, data-scaled intelligence that exhibits five critical properties:

\begin{enumerate}[itemsep=0pt]
\item Learns from synthetic offline data rather than requiring online exploration in operational plants
\item Proposes multiple admissible actuation strategies rather than predicting a single labeled trajectory
\item Is validated through closed-loop simulator execution rather than through parameter-level losses
\item Transfers across reactor types, operating regimes, and monitoring horizons through lightweight adaptation rather than ground-up redesign
\item Achieves reliability improvements through data scaling rather than through architectural complexity or reward engineering
\end{enumerate}

Such a system would fundamentally alter the economics and operational flexibility of nuclear deployment. Operators could adapt a pre-trained foundation model in hours rather than spending months tuning controllers for each new small modular reactor design. They could generate and validate dozens of candidate strategies in real time rather than running computationally expensive model-predictive control solvers during dynamic grid events. The industry could deploy data-driven policies whose tail-risk suppression has been empirically validated across tens of thousands of synthetic scenarios rather than relying on conservative engineering margins to bound brittle, plant-specific controllers.

Treating reactor control as a generative sequence-modeling problem, in which a compact language model writes numeric actions and a high-fidelity simulator acts as the critic, offers an alternative to optimization-based control design for a restricted class of single-step maneuvers in simulation. We do not claim that the present results constitute a paradigm shift; we test whether the approach is viable on a controlled benchmark and report what it does and does not establish.

\begin{figure}[htbp]
  \centering
  \includegraphics[width=1\linewidth]{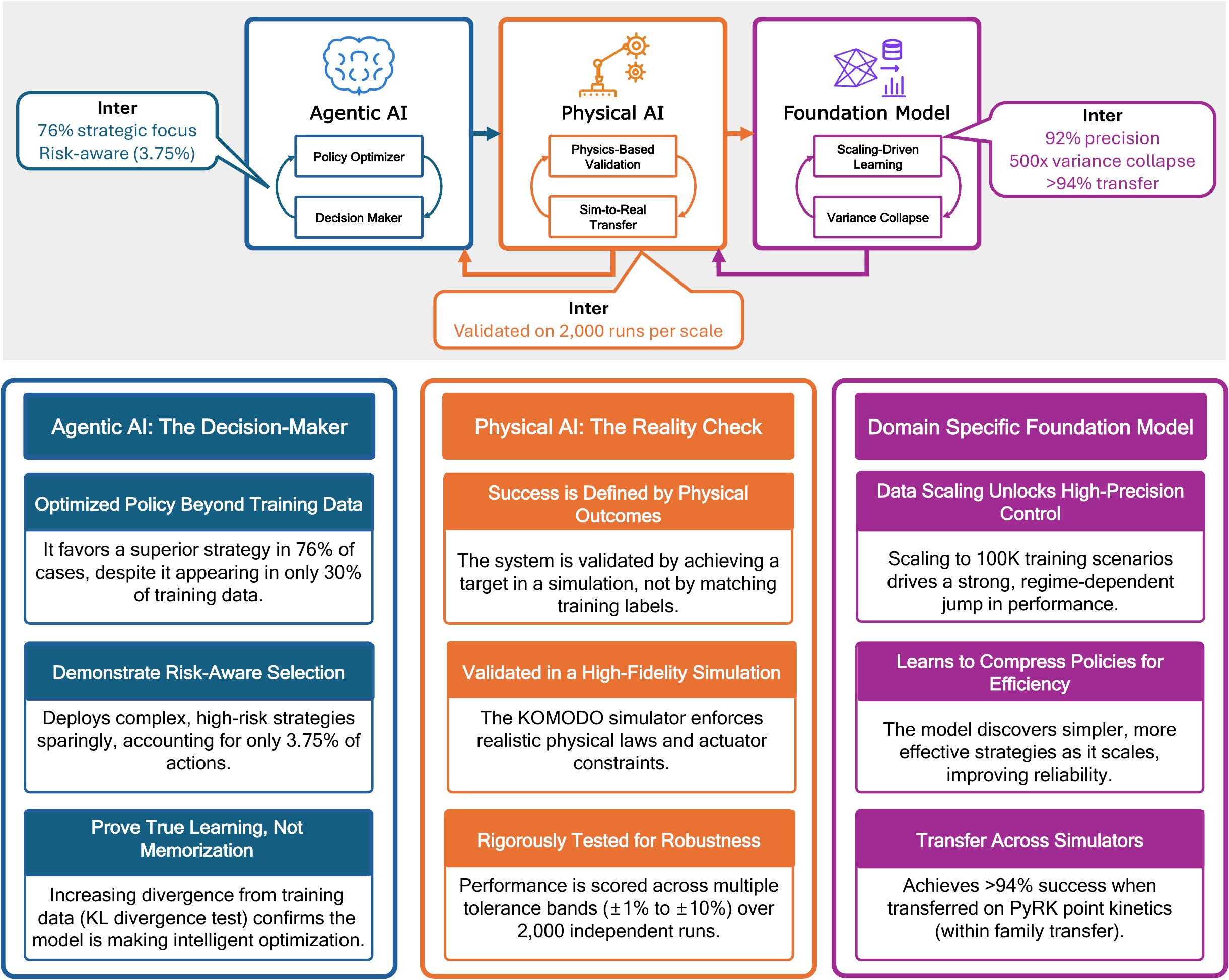}
\caption{\textbf{Integrated framework for Agentic Physical AI in nuclear reactor control.} Three interconnected paradigms form a coherent system: \textbf{(Left) Agentic AI}: the compact 360M-parameter model optimizes runtime policy away from balanced training distribution (KL divergence between runtime and balanced training distribution remains substantial at every scale; Table~\ref{tab:entropy_kl_recomputed}), concentrating 76\% of actions on single\_b2 strategies despite only 30\% training frequency, and deploying brittle multi-bank coordination sparingly (3.75\%). \textbf{(Center) Physical AI}: success is defined by closed-loop execution in KOMODO simulator achieving target power within tolerance bands ($\pm$1--10\%), not by parameter proximity to labels; the model learns to navigate the physics-constrained feasible manifold $\mathcal{M}_{\text{feas}}$ through 2,000 independent validation runs per scale. \textbf{(Right) Foundation Model}: scaling from 1K to 100K scenarios drives strong scaling behavior with regime-dependent gains: sub-1\% precision jumps from 6.0\% to 92.0\%, variance collapses 500×, the runtime policy concentrates on a sub-region of the actuation space at every scale (Table~\ref{tab:entropy_kl_recomputed}), and the model transfers to PyRK point kinetics with >94\% success. The two-phase curriculum (Phase 1: grammar learning via CPT; Phase 2: task conditioning via LoRA) separates domain structure from task specialization, enabling reusable priors. Data scaling provides the feedback loop that stabilizes agentic policies through outcome-centric validation.}
\label{fig:integrated_framework}
\end{figure}

Figure~\ref{fig:integrated_framework} demonstrates this paradigm through three interconnected components. A compact 360M-parameter language model generates candidate actuation sequences, concentrating 76\% of runtime actions on single\_b2 strategies despite only 30\% training frequency—evidence of agentic policy optimization rather than dataset imitation. Closed-loop execution in the KOMODO simulator validates each proposal against physics-based success criteria within $\pm$1--10\% tolerance bands, defining correctness by achieved outcomes rather than parameter proximity. Data scaling from 1K to 100K scenarios drives strong scaling behavior with regime-dependent gains: success at $\pm$1\% tolerance jumps from 6.0\% to 92.0\%, variance collapses 500-fold, and the runtime policy concentrates on a sub-region of the actuation space at every scale (Table~\ref{tab:entropy_kl_recomputed}). The framework transfers to PyRK point kinetics with >94\% success, demonstrating cross-simulator generalization characteristic of domain-specific foundation models.

The emerging language of Agentic AI and Physical AI in robotics and cyber-physical systems provides a useful lens. In robotics, agentic systems are defined not only by perception or prediction, but by the ability to convert high-level intent into sequential action under environmental constraints~\cite{xi2025rise,shavit2023practices,zeng2023large}. Recent work demonstrates that compact vision-language models can function as decision-making agents for autonomous systems~\cite{arora2024g}, while physics-informed neural networks enable rapid physical reasoning for embodied manipulation tasks~\cite{kanwar2026phyplan}. Vision-language-action models and embodied foundation models show that generative sequence modeling can serve as a policy substrate that generalizes across tasks, tools, and settings~\cite{ma2026survey,zhong2025survey,shao2025large}. PaLM-E and RT-2, for example, treat action selection as token generation grounded in physical outcomes and challenge the belief that robust control must always be encoded through task-specific analytic designs~\cite{driess2023palm,zitkovich2023rt}. Nuclear systems differ profoundly from consumer robotics in risk profile and regulatory posture, but the conceptual advance is transferable: a control policy can be learned as a generative model over admissible actuation sequences, provided that physical constraints and rigorous execution-level validation sit at the center of the evaluation protocol.

Three developments make this approach timely and tractable.

First, mature digital-twin ecosystems have eliminated data scarcity. Open-source reactor simulators such as KOMODO~\cite{imron2019development}, RELAP~\cite{relap5d2005manual}, and PyRK~\cite{huff2015pyrk} now provide high-fidelity, deterministic physics engines capable of generating millions of synthetic training scenarios at modest computational cost. Recent advances in operator learning and surrogate modeling~\cite{kobayashi2024deeponet,kobayashi2024improved,hossain2025virtual} 
demonstrate that neural operators can achieve 1,000× speedup over traditional computational methods while maintaining prediction accuracy, enabling real-time digital twin inference without continuous retraining. This eliminates the data scarcity that has historically constrained learning-based approaches in nuclear engineering. The present study leverages KOMODO to construct a balanced corpus of admissible single-bank and two-bank control patterns, demonstrating that structured corpora for offline training are now accessible without access to proprietary plant data.

Second, compact foundation models and parameter-efficient adaptation have made deployment tractable. The emergence of small language models (SmolLM2-360M~\cite{allal2025smollm2}, Phi-3~\cite{abdin2024phi3}, Qwen2.5~\cite{qwen2024qwen25}) and parameter-efficient fine-tuning methods (LoRA~\cite{hu2022lora}, QLoRA~\cite{dettmers2023qlora}) makes it feasible to train deployable models on commodity GPUs in hours rather than requiring supercomputing infrastructure~\cite{zhuang2023survey}. This addresses the computational barrier that has kept foundation-model approaches out of reach for safety-critical domains. Our two-phase curriculum separates structural grammar learning from intent-conditioned actuation generation: the first phase internalizes numeric regularities of valid control commands without exposure to power tokens~\cite{devlin2019bert,spithourakis2018numeracy,wallace2019nlp}, and the second phase grounds these commands in desired power transitions through supervised low-rank adaptation~\cite{bengio2009curriculum,gururangan2020don}. This two-stage design mirrors the separation in many foundation-model pipelines between general-domain structural learning and task-specific grounding~\cite{brown2020language,bommasani2021opportunities}, yet remains computationally light enough to fit within modest hardware budgets.

Third, policy alignment with scientific foundation models creates institutional momentum. The White House Genesis Mission (Executive Order, 24 November 2025) explicitly calls for DOE-led development of AI agents and scientific foundation models for priority areas including nuclear fission and fusion energy~\cite{whitehouse2025genesisEO,whitehouse2025genesisFS}. This institutional signal reflects a broader recognition that the next generation of scientific breakthroughs (in climate, materials, energy, and health) will emerge not from scaling general-purpose models, but from building domain-specific foundation models that combine large diverse corpora with physics-grounded evaluation protocols~\cite{zhang2025artificial}.

Analogous domain-specific foundation models have already transformed other scientific domains. In Earth-system prediction, Aurora shows how a single large-scale model trained on extensive geophysical data can be fine-tuned across tasks such as weather, air quality, ocean waves, and tropical cyclone tracking, often surpassing operational baselines at far lower computational cost~\cite{bodnar2025foundation}. In materials science, generative and operator-style models for inorganic crystals and atomistic simulations demonstrate that large, structured training corpora and physics-aware architectures can accelerate discovery and prediction across composition, temperature, and pressure ranges~\cite{zeni2023mattergen,merchant2023scaling,yang2024mattersim}. In drug discovery, chemical language models have become a prominent framework for de novo molecular generation, linking string-based molecular representations with property-guided design objectives~\cite{grisoni2023chemical}.
 Domain-specific foundation models succeed when they combine large diverse training corpora with domain-structured representations and evaluation protocols that prioritize physically or experimentally grounded outcomes over narrow imitation~\cite{willard2022integrating,zhang2025artificial,bommasani2021opportunities}.

Nuclear reactor control is arguably a more stringent test case: smaller in absolute data volume, richer in constraint structure, and intolerant of failures. If a domain-specific foundation model can succeed here, it establishes a template for safety-critical cyber-physical systems across aerospace, chemical processing, and grid stability~\cite{hassan2024application,tran2019safety}. The convergence of these three factors (synthetic data abundance, compact model tractability, and institutional alignment) creates a unique window to challenge the bespoke-controller paradigm that has dominated nuclear engineering for five decades.

This paper advances an Agentic Physical AI framework that yields a domain-specific foundation model for nuclear reactor power control by integrating three operationally defined principles.

Agentic AI: Policy optimization over admissible strategies. We call a control system agentic when it selects among multiple admissible actuation strategies rather than predicting a single labeled action~\cite{xi2025rise,shavit2023practices}. Our model treats reactor power maneuvering as generative policy selection. Given a desired power transition, the model generates a control vector by sampling from learned distributions over rod positions, time scales, and speeds. The agentic dimension emerges because the model, trained only on success-conditioned offline data, discovers that multiple actuation families (single-bank, simultaneous two-bank, sequential) can all achieve the target, yet at runtime it concentrates probability mass on strategies that maximize closed-loop success. This is not behavioral cloning of a fixed label. It is discovery of a reusable, physically grounded control prior that deviates from the balanced training mixture. The emergent preference for single-bank strategies, despite balanced training data, is evidence that the model is optimizing for reliability rather than imitating surface patterns.

Physical AI: Outcome-centric validation through simulator execution. We use ``Physical AI'' in the sense established in the embodied-AI and robotics literature~\cite{driess2023palm,zitkovich2023rt,xi2025rise}: a learning system whose correctness is adjudicated by executing its outputs in a physics-based environment, rather than by symbolic equation-based constraints embedded in the loss. This usage is distinct from physics-informed machine learning (e.g., physics-informed neural networks, neural operators), in which the governing equations are encoded directly into the architecture or loss as soft or hard constraints. In our setting, physics is enforced \emph{indirectly}: the model proposes a six-parameter rod command, and a high-fidelity simulator integrates the governing neutron-kinetic and thermal-hydraulic equations to determine whether the command achieves the target outcome. The model itself is not equation-constrained; the simulator is. We adopt ``Physical AI'' to emphasize that validation is grounded in physical execution rather than in numerical or perceptual plausibility, and we use ``physics-informed'' or ``equation-constrained'' for loss-level approaches to avoid terminological overlap. Instead of optimizing a parameter-space loss such as mean absolute error on rod positions, we embed the learned policy in a closed-loop validation protocol. Candidate control vectors must (i) satisfy strict actuator limits on rod speeds, positions, and insertion ranges, (ii) be executed deterministically in the KOMODO reactor simulator~\cite{imron2019development}, and (iii) achieve terminal power within specified tolerance bands. Physical realizability, not label proximity, is the success criterion. This tight coupling forces the model to internalize the inverse dynamics of safe reactor control under realistic kinetic and thermal-hydraulic constraints~\cite{johnson2010modeling}. Reliability emerges from the structure of the task: only control proposals that respect physics are counted as successful.

Domain-specific foundation model: Scaling-driven emergence of reusable priors. We define a foundation model as domain-specific when it exhibits scaling-driven emergence of reusable control priors and cross-regime robustness~\cite{bommasani2021opportunities}. In natural language and vision, foundation models combine large diverse corpora, structured representations, and task-agnostic pretraining~\cite{brown2020language,devlin2019bert}. In nuclear reactor control, a more constrained domain, we propose that a domain-specific foundation model is one that demonstrates: (a) qualitative improvement in reliability and tail-risk suppression as data volume increases from one thousand to one hundred thousand scenarios, (b) consistently low-variance performance across operational regimes (small-signal, medium, and large-signal maneuvers), (c) structural transferability across distinct physical engines and variable monitoring windows through a unified learning curriculum~\cite{bengio2009curriculum,gururangan2020don}, and (d) emergent simplification of learned policies that aligns with domain heuristics, indicating that the model has internalized fundamental structure rather than memorizing examples. 

This work establishes scaling-driven emergence in single-step power maneuvers. Extension to multi-step procedures, diverse reactor types, and expanded objectives will complete the foundation-model pathway~\cite{bodnar2025foundation,zeni2023mattergen,merchant2023scaling}.

The agentic AI core generates candidate actuation sequences~\cite{ma2026survey,xi2025rise}, the Physical AI loop filters them through simulator execution~\cite{driess2023palm,zitkovich2023rt}, and data scaling drives the emergence of reusable, domain-grounded priors~\cite{bommasani2021opportunities,allal2025smollm2}. This defines reactor control as a data-centric policy generation problem in which physics validates correctness and data-driven scaling drives robustness, rather than as a purely online optimization problem~\cite{hu2022model,huang2023model,mostafa2024improved}.

This approach challenges three longstanding assumptions in nuclear control: that each plant requires bespoke controller design (we show offline synthetic data enables cross-regime adaptation), that safe generalization requires conservative engineering margins (we demonstrate data-driven policies achieve high reliability without online exploration), and that low parameter error predicts physical success (we prove outcome-centric validation decouples success from label proximity in many-to-one actuation spaces).

The novelty of this perspective becomes clearer when contrasted with mainstream uses of large language models in nuclear engineering. Most current studies are text-centric: document classification, regulatory summarization, or natural-language interfaces layered on diagnostic tools~\cite{al2024scalable,lee2025large,xian2025knowledge,kwon2024sentiment,sun2025youtube,chandra2024exploring,dave2024integrating}. These applications serve the cyber layer of the plant enterprise, but they do not address the creation of low-level numeric actuation proposals that must survive closed-loop physical execution. The present work shifts the focus from language models that talk about nuclear systems to compact physical agents that propose actions for nuclear systems, under rigorously defined execution constraints.

A second axis of novelty is the explicit focus on data scaling as a surrogate for safety-relevant robustness when online rollouts are prohibited. We show that scaling synthetic control corpora from 1,000 to 10,000 and 100,000 scenarios transforms the model from a high-variance approximator into a reliable generator whose tail risk collapses under simulator execution. Mean power error drops from 10.05\% to 0.61\%, severe failures disappear, and success within $\pm$5\% reaches 97.4\% across maneuver regimes, all without online exploration or explicit safety rewards. This scaling-driven transition provides a new lens on reliability-focused learning under the sampled distribution: offline supervised scale can drive the emergence of stable, physically grounded policies. The model never receives an explicit reward signal for reliability, yet it converges toward simpler actuation patterns that suppress catastrophic errors within the sampled distribution, revealing an emergent alignment between data-driven policy selection and practical control-room heuristics.

This approach enables order-of-magnitude improvements in both reliability and operational tempo. Conventional model-predictive control pipelines impose substantial computational costs when operators wish to explore multiple alternatives; the proposed framework generates and screens candidate control vectors in minutes rather than hours, changing how operational decisions can be supported under dynamic grid demands, equipment constraints, or emergency conditions.

\section*{Results}

The Introduction established three structural barriers to learning-enabled nuclear control: (i) the many-to-one structure of reactor actuation, where supervised learning's assumption of unique correct answers penalizes physically equivalent alternatives as variance; (ii) exploration risks that prohibit reinforcement learning's trial-and-error in safety-critical systems; and (iii) the gap between verbal plausibility and execution-level correctness, where models that sound right numerically fail when embedded in physical simulators. Here, we demonstrate that a compact language model trained on offline synthetic data and validated through closed-loop simulator execution can circumvent all three barriers simultaneously. The model does not predict single labeled trajectories. It generates diverse admissible proposals from a learned manifold, addressing the many-to-one problem (Agentic AI). It does not optimize parameter-space losses. Success is defined by physical outcomes under deterministic simulator execution, addressing the execution-correctness gap (Physical AI). And it does not rely on online exploration or architectural scaling. Reliability emerges from data scaling alone, establishing early steps toward a domain-specific foundation model for nuclear control (toward domain-specific foundation models).

\subsection*{Experimental setup and validation protocol}

\begin{figure*}[t]
\centering
\includegraphics[width=0.95\linewidth]{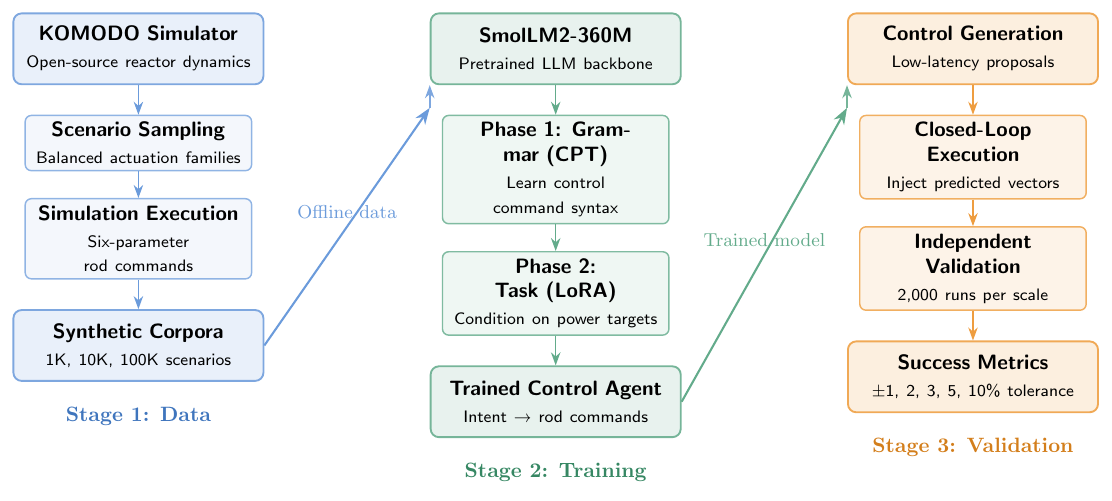}
\vspace{+2mm}
\caption{\textbf{Experimental workflow for Agentic Physical AI in nuclear reactor control.} The pipeline consists of three stages: \textbf{(Stage 1) Data generation}: KOMODO simulator generates synthetic corpora at three scales (1K, 10K, 100K) with balanced actuation families (single-bank, simultaneous, sequential) to prevent trivial dataset bias. \textbf{(Stage 2) Two-phase curriculum training}: SmolLM2-360M backbone undergoes Phase 1 continued pretraining (CPT) to learn control command grammar without power conditioning, followed by Phase 2 supervised LoRA fine-tuning to map power targets to six-parameter rod commands. \textbf{(Stage 3) Physical AI validation}: trained models generate control proposals validated through 2,000 independent closed-loop KOMODO simulations per scale, with success measured by terminal power accuracy across five tolerance bands ($\pm$1, 2, 3, 5, 10\%), not by parameter proximity to labels. This outcome-centric evaluation enables agentic behavior by allowing the model to select among multiple admissible solutions within the physics-constrained feasible manifold.}
\label{fig:komodo-workflow}
\vspace{-4mm}
\end{figure*}

To evaluate whether a compact language model can reliably generate physically meaningful reactor control actions within an Agentic Physical AI framework, we trained SmolLM2-360M~\cite{allal2025smollm2} using a structured, two-phase curriculum designed to separate numeric grammar acquisition from task conditioning. The overarching question is whether a small, instruction-tuned language model can act as the core of a toward domain-specific foundation model for nuclear control: one that learns a stable control manifold from data and then deploys it agentically in closed-loop physical simulation. Figure~\ref{fig:komodo-workflow} summarizes the full pipeline, including dataset construction, curriculum design, and closed-loop validation in a high-fidelity reactor simulator.

The model is built on a pretrained language-model backbone (SmolLM2-360M) and operates entirely on tokenized text. Each scenario is serialized into a numeric string under a fixed schema (initial power, target power, six rod parameters), and both training and inference proceed at the token level. The term ``grammar-based prior'' used in this paper refers to the syntactic and structural regularity of valid control-command strings learned in Phase~1: which numeric ranges, orderings, and parameter combinations correspond to executable commands. We do not introduce a separate symbolic grammar or rule system; rather, the standard language-model objective is applied to the numeric serialization, and the structural regularity emerges from exposure to the training data. The data are therefore numeric in semantic content but discrete-token in representation. Continuous side information is introduced through a small projection adapter that maps a real-valued input into the token-embedding space, jointly trained with a low-rank update to the backbone.

We generated large-scale synthetic datasets containing 1,000, 10,000, and 100,000 single-step control scenarios using the open-source KOMODO simulator~\cite{imron2019development}, a validated pressurized-water-reactor (PWR) dynamics code.

Each data sample comprises an eight-dimensional vector encoding (i) the initial reactor power, (ii) the target power, and (iii) a six-parameter control-rod command:
\begin{equation}
(\texttt{b1\_pos}, \texttt{b1\_time}, \texttt{b1\_speed}, \texttt{b2\_pos}, \texttt{b2\_time}, \texttt{b2\_speed}).
\end{equation}

The initial power was normalized to 1.0, and the target power was expressed as a fractional deviation. Bank positions were expressed in simulator steps (0 to 180), time in seconds, and speed in steps per second. Before each scenario, bank 1 and bank 2 were initialized at 180 and 100, respectively. These correspond to typical PWR operating configurations in which coarse and fine-control banks are differentially positioned to maintain reactivity margin. As discussed in the Discussion section, these initialization patterns exert a measurable inductive influence on the solution space explored by the model and contribute to the emergent policy structure of the Agentic Physical AI.

The model was trained using a two-phase curriculum that separates structure learning from task grounding. In Phase 1 (grammar learning via continued pretraining), power inputs were removed and the model predicted only the six control-rod parameters. This phase acquires numeric grammar and structural priors over the control space independent of downstream objectives. The model learns an approximate representation of the manifold $\mathcal{M}_{\text{grammar}} \subset \mathbb{R}^6$ of valid control commands, defined by simulator constraints (positions in [0, 180], speeds $>0$, times $\geq 0$), physical feasibility (no instantaneous motion, achievable accelerations), and operational norms. Exposure to 1K, 10K, or 100K valid samples teaches the manifold's shape and structure, creating a prior inherited by Phase 2.

In Phase 2 (task conditioning via supervised LoRA fine-tuning), the full eight-number vector (initial power, target power, six control parameters) was reinstated. Low-rank adaptation (LoRA) layers condition the pretrained grammar on mapping power-demand changes to control vectors, with separate fine-tuning on each dataset scale to quantify scaling effects. LoRA provides three methodological advantages: it preserves the Phase 1 grammar prior by constraining adaptations to a low-rank subspace, preventing catastrophic forgetting; it isolates data-volume effects by fixing backbone capacity while varying only the LoRA adaptation at each scale; and it enables compression by forcing the model to discover high-utility strategies rather than memorizing combinations. The Direct LoRA baseline (skipping Phase 1 entirely) achieved near 0\% success with mean error exceeding 100\%, validating the necessity of the two-phase curriculum.

All model variants were evaluated using 2,000 independent, out-of-sample simulation runs in KOMODO~\cite{imron2019development}. Each evaluation initialized the simulator with fixed bank positions and nominal thermal-hydraulic state, injected a model-generated six-parameter control vector without modification, propagated reactor dynamics to a fixed evaluation horizon, and extracted achieved power for comparison to the target. Each scenario was scored across five tolerance bands ($\pm$1, 2, 3, 5, and 10\%), with $\pm$5\% as the primary success criterion following common practice in transient power maneuvering and $\pm$10\% assessing robustness under large deviations.

Critically, success is defined not by parameter-space proximity but by closed-loop physical outcome: this is the defining feature of Physical AI validation. Define the forward reactor dynamics as $\mathcal{F}: \mathcal{A} \to \mathcal{O}$, mapping actions to outcomes. Classical supervised learning optimizes:
\begin{equation}
\mathcal{L}_{\text{parameter}} = \mathbb{E}\left[\|\mathbf{a}_{\text{label}} - \mathbf{a}_{\text{predicted}}\|_2^2\right].
\end{equation}
Physical AI instead evaluates:
\begin{equation}
\mathcal{L}_{\text{outcome}} = \mathbb{I}\left[\mathcal{F}(\mathbf{a}_{\text{predicted}}) \notin \mathcal{B}_\delta(o^*)\right],
\end{equation}
where $\mathcal{B}_\delta(o^*)$ is the tolerance band around the target outcome (for example, $\pm$5\% power). This decoupling enables agentic behavior: the model is free to select among the manifold $\mathcal{M}_{\text{sol}}(o^*)$ of admissible strategies, provided they achieve physical success. A control proposal that perfectly matches a label's rod parameters but produces the wrong terminal power is deemed a failure. Conversely, a proposal that deviates substantially from any single labeled example but achieves the target power is a success.

The evaluation protocol operationalizes agentic behavior through three design choices addressing the many-to-one barrier. The model trains on balanced actuation families (30/30/30/10 distribution across single-bank and multi-bank strategies) yet concentrates runtime execution on strategies it discovers to be most reliable (76.1\% single\_b2 usage despite only 30\% training frequency). Success is defined by achieving the physical objective (terminal power within tolerance), not matching labeled parameters; a control vector differing substantially from training examples yet achieving the target is scored successful. The model navigates a manifold of admissible solutions in PWR control, where multiple rod sequences achieve identical power outcomes, by deploying consistent, outcome-optimized strategies rather than imitating single trajectories. These properties distinguish agentic policy formation from passive imitation.

To assess how difficulty varies with operational challenge, all validation cases were grouped into three bins: small shifts ($<10\%$ power change, fine maneuvering), medium shifts (10--30\%, routine load-following), and large shifts (30--50\%, aggressive or emergency control extending beyond standard ramp-rate envelopes). This stratification distinguishes tasks sensitive to rod-speed scheduling (small shifts) from those dominated by reactivity insertion magnitude (large shifts). Small shifts demand fine-grained control precision; large shifts test the ability to compute large reactivity changes. Uniform success across all three regimes indicates internalization of the multiscale structure of reactor control, a characteristic of foundation-model behavior in physical systems.

Every simulation run was executed in isolation, with no carry-over state or memory between cases. Temporary files and simulator caches were removed after each run, ensuring that only the achieved power and metadata were retained. Every validation sequence therefore constitutes a clean, closed-loop evaluation under fully deterministic conditions. This strict reproducibility protocol ensures that observed improvements in success rates are genuine manifestations of model capacity rather than simulator artifacts or initialization biases.

Together, these elements establish one of the first large-scale, fully automated evaluation pipelines for language-model-based reactor controllers in closed-loop physical simulation. The protocol is explicitly designed to test whether a compact language model can function as an Agentic Physical AI and to provide early evidence that it is moving toward domain-specific foundation-model behavior for nuclear control.

\subsection*{Offline parameter test on labeled controls}

Before assessing closed-loop behavior in the reactor simulator, we first quantified the model's ability to reproduce labeled six-parameter control vectors in an offline setting. This experiment isolates the model's representational accuracy from the nonlinear reactor dynamics and provides a lower-bound estimate of how well the model internalizes the structure of feasible actuation commands. Offline evaluation was conducted by comparing each model's predicted control vector against ground-truth labels using mean absolute error (MAE). Table~\ref{tab:offline_mae_comparison} summarizes results for the three dataset scales.

To ensure statistically meaningful comparisons, the 1K-trained model was evaluated on 100 held-out samples, the 10K model on 1,000 samples, and the 100K model on 10,000 samples. Across dataset scales, the MAE for \texttt{b1\_pos} remained in the 20 to 30 step range for most control families. The notable exception is the \texttt{single\_b2} case, where errors were markedly lower. This pattern indicates that the model learns a more compact mapping when only one bank is active, and that the \texttt{single\_b2} configuration forms a particularly stable attractor in the learned control manifold. As we discuss later, this aligns with the model's emergent preference for \texttt{single\_b2} solutions in closed-loop experiments, suggesting that both the dataset distribution and initial-rod configurations embed inductive biases that shape the learned policy landscape.

For \texttt{b2\_pos}, errors were consistently smaller in the \texttt{single\_b1} and \texttt{single\_b2} conditions compared to multi-bank cases. This asymmetry highlights an important structural characteristic of the dataset: control scenarios in which one bank is dominant present a simpler, more separable mapping that the model captures more accurately. Multi-bank cases, by contrast, require reasoning over coupled time, position, and speed relationships between two actuators, a substantially harder task in the offline regime where no physical feedback is available.

Temporal and speed parameters exhibited similar trends. Across all dataset scales, MAEs for time and speed were smaller for bank 2 than for bank 1. One plausible explanation is that the initial-condition asymmetry, with bank 1 fully withdrawn and bank 2 partially inserted, reduces the effective action space for bank 2, making its temporal and kinematic fields easier to predict. Conversely, bank 1 actions span a wider feasible range, enabling more diverse training examples but increasing reconstruction difficulty. These inductive differences propagate through Phase 1 grammar learning, ultimately shaping the model's internal representation of physically plausible control actions.

Offline MAE values do not predict closed-loop performance. Offline tests evaluate numerical consistency with labeled data, while closed-loop control requires reasoning about the reactor's nonlinear response to actuation. The 100K model's strong closed-loop performance (97.4\% success) despite modest offline MAE (26.95 steps for b1\_pos) exemplifies Physical AI validation, where execution success decouples from parameter proximity.

This divergence addresses the third barrier from the Introduction: the gap between verbal (or numerical) plausibility and execution-level correctness. Classical supervised learning would predict that high offline MAE correlates with poor closed-loop success. Instead, the 100K model achieves 97.4\% closed-loop success despite moderate parameter-level errors (for example, b1\_pos MAE of 26.95 steps for single\_b1 cases). 

This decoupling occurs because the model has learned to reason about forward dynamics. Rather than memorizing inverse parameter mappings, the model has learned the structure of the solution manifold. In many-to-one control spaces, multiple distinct parameter combinations map to the same physical outcome. Let $\mathcal{M}_{\text{sol}}(o^*)$ denote the set of all actions that achieve target outcome $o^*$:
\begin{equation}
\mathcal{M}_{\text{sol}}(o^*) = \{\mathbf{a} \in \mathcal{A} : \mathcal{F}(\mathbf{a}) = o^*\}.
\end{equation}
A model optimized for parameter proximity is penalized for deviating from a specific labeled example, even when that deviation leads to an equally valid or superior solution. A model validated through closed-loop execution is free to explore the solution manifold and concentrate on high-reliability strategies. 

This is Physical AI: correctness is defined by what happens when the reactor executes the command, not by how closely the command matches a label. An agentic controller is not judged by how closely it imitates any single reference; it is judged by whether its chosen action, drawn from the learned distribution over admissible strategies, produces the correct physical outcome.

The offline analyses demonstrate that the model learns numerical grammar and relative relationships among rod parameters, and that scaling smooths the learned parameter landscape before simulator interaction. This behavior, which internalizes stable numerical structure for agentic deployment under physical constraints, characterizes domain-specific foundation models.

\begin{table}[htbp]
\centering
\caption{\textbf{Offline per-field MAE by model scale and scenario type}}
\label{tab:offline_mae_comparison}
\resizebox{\textwidth}{!}{%
\begin{tabular}{lrrrrrrrr}
\toprule
\textbf{Scenario} & \textbf{Model} & \textbf{n} & \textbf{b1\_pos} & \textbf{b1\_time} & \textbf{b1\_speed} & \textbf{b2\_pos} & \textbf{b2\_time} & \textbf{b2\_speed} \\
\midrule
single\_b1 & 1K & 27 & 26.333 & 6.007 & 0.944 & 6.444 & 5.011 & 0.430 \\
           & 10K & 301 & 27.375 & 5.050 & 0.872 & 4.963 & 3.712 & 0.428 \\
           & 100K & 3,009 & 26.954 & 5.378 & 1.020 & 5.143 & 3.204 & 0.493 \\
\addlinespace
single\_b2 & 1K & 31 & 4.226 & 0.871 & 0.119 & 10.774 & 4.745 & 0.497 \\
           & 10K & 303 & 3.089 & 1.300 & 0.184 & 6.601 & 4.857 & 0.546 \\
           & 100K & 3,006 & 2.521 & 1.091 & 0.229 & 6.673 & 4.033 & 0.617 \\
\addlinespace
simultaneous & 1K & 31 & 20.194 & 6.342 & 0.977 & 11.806 & 5.368 & 0.694 \\
             & 10K & 296 & 27.774 & 5.774 & 1.117 & 14.196 & 5.479 & 0.648 \\
             & 100K & 2,992 & 30.224 & 5.708 & 1.146 & 15.252 & 4.842 & 0.741 \\
\addlinespace
sequential & 1K & 11 & 48.273 & 7.155 & 1.627 & 19.545 & 6.464 & 0.682 \\
           & 10K & 100 & 39.620 & 7.415 & 1.227 & 18.310 & 6.032 & 0.633 \\
           & 100K & 993 & 37.288 & 7.569 & 1.253 & 17.296 & 5.151 & 0.741 \\
\bottomrule
\end{tabular}%
}
\end{table}

\subsection*{Effect of data scaling (1K, 10K, and 100K) on accuracy, stability, and tail-risk collapse}

Scaling the training corpus from 1K to 100K reactor control scenarios produces qualitative changes in policy reliability characteristic of emergent policy formation in domain-specific foundation models. Foundation models (across language, vision, climate, and materials) transition from brittle pattern matching to stable, geometry-aware reasoning through data scaling. We observe an analogous transition in accuracy, robustness, and tail behaviour.

To characterize the data-scaling behavior, we examine whether the success rate $S(n,\delta)$, defined as the proportion of 2,000 runs achieving terminal power within tolerance $\delta$, exhibits power-law behavior as a function of dataset size $n$.

\begin{figure}[htbp]
  \centering
  \includegraphics[width=\linewidth]{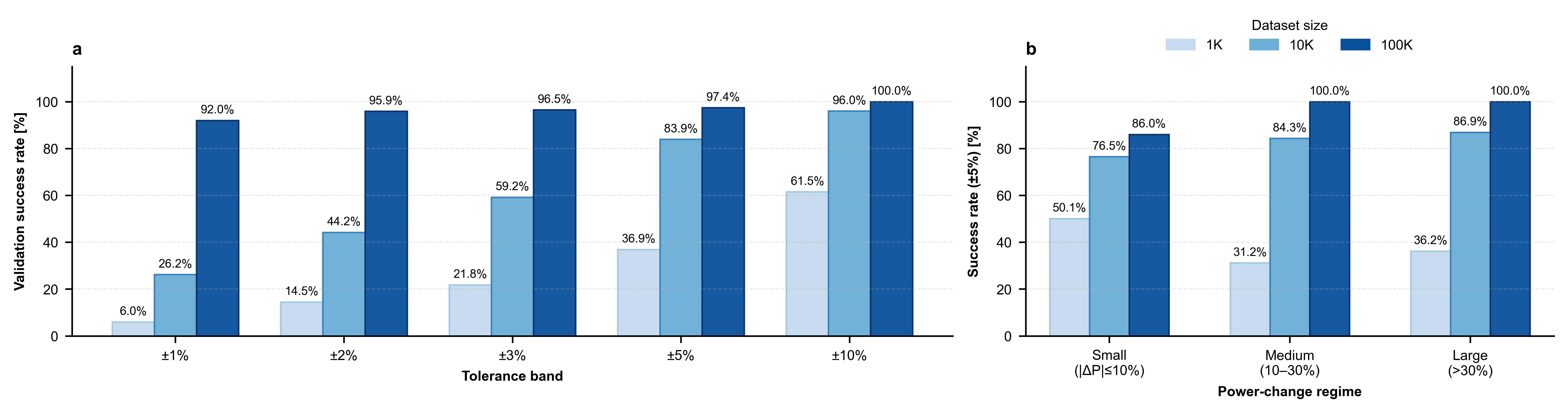}
\caption{\textbf{Scaling of validation success and regime robustness with dataset size.}
  \textbf{a.} Validation success rates across tolerance bands show a sharp improvement between 10K and 100K, revealing the emergence of a stable, high-precision control policy, with sub-1\% accuracy jumping from 26.2 to 92\%.
  \textbf{b.} Performance stratified by power-change bins shows that the 100K model achieves regime-consistent precision that is absent in the 1K and 10K variants.}
  \label{fig:komodo-tolerance}
\end{figure}

Figure~\ref{fig:komodo-tolerance}a summarizes validation success across tolerance bands for the 1K, 10K, and 100K models. Define the critical tolerance as 5\%, the primary success band. The observed success rates at this threshold are:
\begin{equation}
\begin{aligned}
S_1 &= 36.9\text{\%}, \\
S_{10} &= 83.9\text{\%}, \\
S_{100} &= 97.4\text{\%},
\end{aligned}
\end{equation}
where subscripts denote dataset size in thousands. The scaling is steep:
\begin{equation}
\begin{aligned}
\Delta S_{1 \to 10} &= S_{10} - S_1 = 47.0\text{\%}, \\
\Delta S_{10 \to 100} &= S_{100} - S_{10} = 13.5\text{\%}.
\end{aligned}
\end{equation}

Define the scaling exponent
\begin{equation}
\alpha(\delta) = \frac{\log_{10}\!\left(S_{100}(\delta)/S_{10}(\delta)\right)}{\log_{10}(10)} = \log_{10}\!\left(\frac{S_{100}(\delta)}{S_{10}(\delta)}\right).
\end{equation}
At the operational $\pm 5\%$ band, $\alpha(5\%) = \log_{10}(97.4/83.9) \approx 0.065$, indicating saturation: the 10K model already captures most of the global structure of the control map. At the strict $\pm 1\%$ tolerance,
\begin{equation}
S_1^{(1\%)} = 6.0\%,\qquad S_{10}^{(1\%)} = 26.2\%,\qquad S_{100}^{(1\%)} = 92.0\%,
\end{equation}
yielding $\alpha(1\%) = \log_{10}(92.0/26.2) \approx 0.55$. The exponent is sub-linear in the formal sense ($\alpha < 1$). We describe the effect as \emph{strong scaling behavior with regime-dependent gains}: the absolute increase at the strict tolerance ($+65.8$ percentage points between $10^4$ and $10^5$ scenarios, a $\sim 3.5\times$ multiplicative gain), the concomitant $\sim 500\times$ variance collapse, and the elimination of $>10\%$ excursions are operationally substantial improvements along a smooth scaling curve. The 100K model resolves the local structure of the control map more sharply than the 10K model.

\begin{figure}[htbp]
  \centering
  \includegraphics[width=\linewidth]{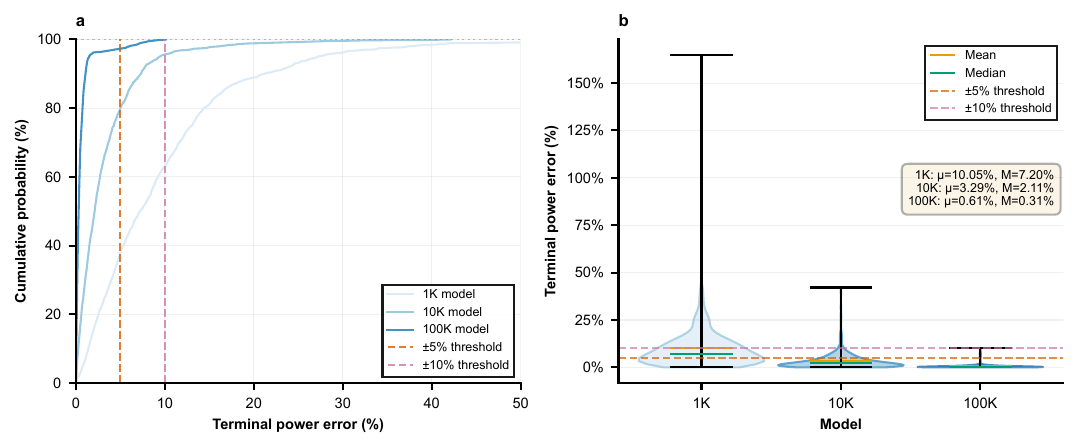}
  \caption{\textbf{Terminal power error distributions across dataset scales}. \textbf{a.} CDF curves highlight tail-risk collapse at 100K scale. \textbf{b.} Violin plots demonstrate narrowing uncertainty and emergence of a stable, low-variance control policy.}
  \label{fig:komodo-combined}
\end{figure}

Under the primary $\pm$5\% criterion, the 1K model succeeded in only 737 of 2,000 runs (36.9\%), indicating failure to learn a coherent mapping from power targets to rod trajectories. At $\pm$10\%, success rose to 61.5\%, but the high variance and wide distribution of errors (discussed later in Figure~\ref{fig:komodo-combined}) demonstrate that the 1K model operates as a high-uncertainty policy generator rather than a meaningful controller. This is consistent with a model that has memorized surface-level trends without internalizing the structure of the underlying control manifold.

The 10K model marks a clear shift in behavior. Success at $\pm$5\% rises to 83.9\%, and at $\pm$10\% rises to 96.0\%. The model now captures the global relationship between demanded power change and required reactivity insertion, producing consistent control vectors across a broad range of targets. However, its low performance in the strictest band ($\pm$1\% at 26.2\%) indicates that while the model has acquired a workable approximation to the control map, it has not yet learned the fine-grained timing and rod-velocity coupling required to navigate the reactor's nonlinearities in the near-equilibrium regime.

A qualitatively different regime emerges at 100K. The 100K model succeeds in 1,947 of 2,000 runs within $\pm$5\% (97.4\%) and achieves 2,000 of 2,000 successes at $\pm$10\%. At $\pm$1\%, the success rate reaches 92\%—a more than threefold improvement over the 10K model and over fifteenfold improvement over 1K. This gain reflects the emergence of a stable internal representation of reactor control dynamics, not incremental smoothing. At this scale, the model has discovered latent structures, a learned manifold of physically feasible control strategies, that enables it to generalize to unseen power demands with remarkable precision. The scaling from 10K to 100K marks the regime in which the model's learned representation shifts from a shallow, task-specific mapping to a deep, physics-aware control prior.

At this scale, the model begins to approximate not only the global control manifold but also its local curvature, capturing how small perturbations in timing, rod position, and rod speed nonlinearly influence terminal power. This local curvature approximation is characteristic of foundation models: they learn not just average behaviors but the higher-order structure of the domain.

This scaling behavior parallels observations from foundation models in other physical domains, where sufficient data diversity and distributional richness drives a shift from memorization to rule abstraction. In the context of nuclear reactor control, this shift becomes evident between 10K and 100K scenarios. Only at the largest scale does the model behave as a reliable agent capable of generating consistent, low-variance actuation patterns across the entire operating envelope.

Figure~\ref{fig:komodo-tolerance}b further dissects the results by stratifying validation scenarios into small, medium, and large power-change bins. This analysis is essential for understanding how the model reasons about different control regimes. Critically, a model that exhibits regime-consistent precision, that is, high success across all power-change magnitudes, has internalized multiscale structure, which is a prerequisite for foundation-model behavior in control.

The 1K model's uneven performance (50.1\% small, 31.2\% medium, 36.2\% large) suggests that it learned only superficial patterns near the equilibrium point and could not extrapolate to larger maneuvers. The 10K model displays a monotonic trend with success rates of 76.5, 84.3, and 86.9\%, respectively, capturing the global scaling between control magnitude and power shift but struggling in the most precise regime where reactor physics is steepest and timing inaccuracies are amplified.

The 100K model behaves qualitatively differently. It achieves 86\% success for small adjustments and 100\% success for both medium and large ranges. This is a defining signature of a domain-specific foundation model: the model exhibits consistent, high success across regimes that tax different aspects of control reasoning (fine timing for small shifts, magnitude for medium shifts, and large reactivity insertion for large shifts). A model that fails in any regime has learned brittle, regime-specific heuristics. A model that succeeds uniformly has internalized the underlying principles of reactor control. These results indicate that large-scale data endow the model with a physically structured internal representation, enabling it to generate rod trajectories that satisfy the strict timing, sensitivity, and reactivity insertion constraints of both small-signal and large-signal regimes. In other words, the model begins to function as a domain-specific foundation model for control, exhibiting consistent performance across an operationally diverse range of applications.

Tail-risk behavior is critical in safety-critical domains. Even a small number of large deviations from the target can trigger protective actions or lead to operational instability. Figure~\ref{fig:komodo-combined} shows the cumulative distribution function (CDF) of terminal power error (panel a) and the corresponding violin plots of error distributions (panel b).

Let $E_i$ denote the terminal power error for run $i$. The tail-risk metric is commonly quantified by the loss in the upper quantiles. Define the $q$-th quantile as:
\begin{equation}
Q_q = \inf\{\epsilon : \Pr[E \le \epsilon] \ge q\}.
\end{equation}
For safety-critical systems, we care about the far tail: the 95th and 99th percentiles. At the 1K scale, $Q_{95}(E) \approx 40$\%, indicating that 5\% of runs exceed 40\% error. At the 100K scale, $Q_{95}(E) \approx 1$\%, representing a 40-fold reduction in tail risk. More dramatically, the worst-case error collapses from approximately 50\% at 1K to under 2\% at 100K.

Formally, define the variance-collapse ratio as:
\begin{equation}
\mathcal{V}_{\text{collapse}} =
\frac{\text{Var}(E)_{1K}}{\text{Var}(E)_{100K}}.
\end{equation}
From the data, $\text{Var}(E)_{1K} \approx 250$ (percent-squared) and $\text{Var}(E)_{100K} \approx 0.5$ (percent-squared), yielding:
\begin{equation}
\mathcal{V}_{\text{collapse}} \approx 500.
\end{equation}
This is a substantial reduction in tail risk, consistent with strong scaling. The model has learned not just to minimize mean error but to minimize variance, to concentrate its probability mass on a tight, reliable band of outcomes.

The 1K model exhibits slow CDF growth and long, heavy tails: only 36.9\% of runs fall within $\pm$5\% and 61.5\% within $\pm$10\%. The violin plot confirms a broad, elongated distribution with high variance, characteristic of an unstable agent whose decisions are sensitive to small variations in input.

The 10K model collapses much of this tail behavior, achieving 83.9 and 96.0\% success under the same thresholds. Its violin plot narrows considerably, yet still shows noticeable spread in the small-error regime. The model understands the broad geometry of the control landscape but lacks the precision necessary for universal reliability.

The 100K model nearly eliminates tail errors entirely. The CDF rises steeply, with 97.4\% within $\pm$5\% and 100\% within $\pm$10\%. The error distribution becomes sharply peaked at zero, with mean and median errors of 0.61 and 0.31\%, respectively. This collapse of variance is not merely an improvement in average accuracy; it is a systemic reduction in policy unpredictability, an essential criterion for deploying AI agents in nuclear environments. The model has learned a tight, reproducible strategy rather than a diffuse, probabilistic one. In information-theoretic terms, the model has achieved high entropy reduction in the space of feasible control outcomes: given a power demand, the model now generates proposals that cluster tightly around a reliable solution.

Data scaling transforms the model from a high-variance approximator into a low-variance policy generator on the sampled distribution. We characterize three empirical regimes: at 1K, success rates are below 50\% across all actuation classes (Figure~\ref{fig:komodo-scenario}a); at 10K, single-bank strategies become consistently reliable but multi-bank coordination remains brittle (Figure~\ref{fig:komodo-scenario}b); at 100K, severe failures on the sampled distribution are eliminated and multi-bank coordination is deployed sparingly (Figure~\ref{fig:komodo-scenario}c, Figure~\ref{fig:komodo-failure}c). We refrain from labeling these regimes as memorization, proto-policy formation, and mature agentic behavior, since the present three data points cannot adjudicate the underlying mechanism.

Scaling enables the model to infer long-range couplings between rod position, timing, and reactivity; manifest low-variance decision policies essential for safety-critical settings; generalize smoothly across power-change regimes; and behave consistently as a unified agent rather than a collection of memorized patterns. The model does not achieve agentic behavior through explicit reward signals, lookahead search, or multi-step planning. Agentic behavior emerges when the model develops sufficient internal structure to reliably optimize outcomes within the physical environment's constraints.

This work establishes a first step, not the complete pathway, toward a domain-specific foundation model for nuclear control. Foundation models are characterized by three properties, all of which we demonstrate in limited scope: 

First, scaling-driven emergence, where quantitative data increases induce strong scaling behavior with regime-dependent gains in capability. We demonstrate: 1K to 10K to 100K drives 36.9 to 83.9 to 97.4\% success, 500-fold variance collapse, elimination of all catastrophic failures, and sub-1\% precision jumping from 6.0 to 26.2 to 92.0\%.

Second, cross-domain transfer, where a single architecture adapts to diverse tasks and environments. Partially demonstrated: the model transfers to PyRK point kinetics with the same two-phase curriculum and zero architectural modification.

Third, reusable structural priors, where learned representations compress high-dimensional action spaces into stable, physically grounded policies. Demonstrated: the model develops an emergent single-bank preference aligning with operator heuristics despite balanced training, with the model concentrating runtime execution on a small subset of admissible strategies (76\% \texttt{single\_b2} at 100K) far from the balanced 30/30/30/10 training mixture (Table~\ref{tab:entropy_kl_recomputed}).

This work establishes enabling conditions within a defined scope: single-step PWR power maneuvers under fixed initial conditions (Bank 1 at 180, Bank 2 at 100). Extension to multi-step procedures, diverse reactor types (BWR, SMR, liquid-salt), additional objectives (xenon override, transient mitigation), and uncertainty quantification defines the complete pathway. We demonstrate the mechanism—scaling plus curriculum plus Physical AI validation—but not yet the full operational envelope.

\subsection*{Comparative advantage over classical control baselines}
\label{sec:baselines}
 \label{adv}

\begin{figure}[t]
  \centering
  \includegraphics[width=\linewidth]{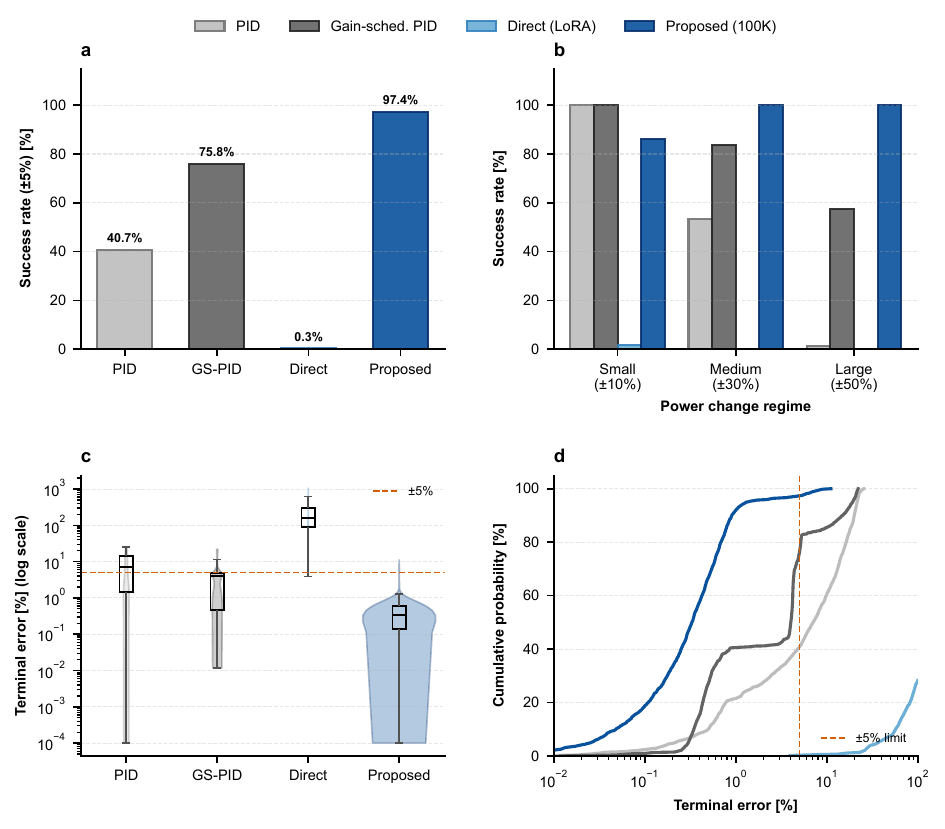}
  \caption{\textbf{Benchmarking Agentic AI against classical and direct-learning baselines.} 
\textbf{a.} Overall success rates ($\pm 5\%$ tolerance) for four models: single-gain PID 
(40.7\%), gain-scheduled PID with six regime-dependent gains (75.8\%), Direct (LoRA) 
ablation (0.3\%), and the Proposed 100K model (97.4\%).
\textbf{b.} Success rates stratified by power-change magnitude. The single-gain PID 
collapses in the large regime due to actuator saturation; the gain-scheduled variant 
recovers most of the gap except in the large-decrease regime; the Proposed model 
maintains high reliability across all regimes.
\textbf{c.} Error distribution (log scale).
\textbf{d.} Cumulative distribution function (CDF) of errors, highlighting the 
foundation model's tail-risk management.}
  \label{fig:comparison}
\end{figure}

To contextualize the performance of the proposed Agentic AI, we benchmarked it against three baselines: a single-gain proportional controller operating predominantly through Bank 2; a gain-scheduled variant of the same controller with six regime-dependent proportional gains; and a Direct (LoRA) baseline in which the foundation-model backbone was fine-tuned directly on the target objective, skipping the Phase~1 grammar-learning stage of our two-phase curriculum. We emphasize that the proportional controller used here is not representative of state-of-the-art classical control practice for reactor power tracking. Gain-scheduled PID, model-predictive control (MPC) with neural surrogates~\cite{hu2022model,xiao2022neural,yin2023design}, sparse-identification-based controllers (SINDy-RL,~\cite{kaiser2018sparse}), and reinforcement-learning baselines for load-following and microreactor control~\cite{gong2024possibilities,nguyen2024reinforcement,tunkle2025nuclear,radaideh2026multistep} routinely outperform single-gain proportional designs in this class of problem. The purpose of the present comparison is therefore narrower: to confirm that the learned policy is non-trivial and that it substantially exceeds a minimally tuned feedback baseline, and to isolate the contribution of the two-phase curriculum via the Direct LoRA ablation. A controlled comparison of the proposed Agentic AI against gain-scheduled MPC, SINDy-RL, and conservative-Q-learning offline-RL baselines on identical KOMODO scenarios is part of ongoing work and will be reported in a separate follow-up study. Figure~\ref{fig:comparison} summarizes the results of this comparative analysis.

KOMODO operates as a single-shot transient solver: a six-parameter rod command is supplied once, the simulator integrates the full reactor transient, and a terminal power is returned. There is no exposed plant state during the transient, and therefore no possibility of true closed-loop feedback control (continuous error integration, derivative damping, or online regulator updates) within a single run. The proposed Agentic AI emits one six-parameter command per scenario; to keep the comparison fair, both classical baselines below are also one-shot mappings from $(P_{\text{initial}}, P_{\text{target}})$ to a six-parameter rod command. The ``single-gain PID'' label, used throughout this section and in Figure~\ref{fig:comparison} for continuity with prior nomenclature, refers in this paper to a calibrated single-shot proportional mapping $\Delta s = K_p\,(P_{\text{target}} - P_{\text{initial}})$ rather than to a closed-loop PID regulator with non-zero integral and derivative gains. The gain-scheduled variant inherits the same one-shot structure with six regime-dependent gains. The term ``closed-loop'' is used throughout this paper in the outcome-validation sense (model outputs are executed in the physics-based simulator and judged by terminal power) rather than in the sense of continuous online feedback regulation.

As shown in Figure~\ref{fig:comparison}a.
Under the $\pm$5\% tolerance criterion, the proposed 100K model achieved a dominant success rate of 97.4\%, whereas the single-gain PID baseline reached only 40.7\%. The Direct (LoRA) baseline failed to learn a meaningful control policy, converging to near 0\% success rate with a mean error of 235\%. This directly supports the claim that a carefully structured curriculum is essential for turning a compact language model into an Agentic Physical AI rather than a brittle function approximator.

The proportional controller's limited success in this comparison reflects two features of the baseline: it is a single-gain design with no scheduling or feedforward term, and it is calibrated on only a small set of symmetric scenarios. A gain-scheduled or model-based controller would be expected to perform substantially better, and we therefore do not claim that the present results establish dominance over modern classical control. What the comparison does demonstrate is that the learned policy is not trivially obtained from minimal feedback tuning, and that it generalizes across power-change regimes without per-regime hand-tuning. The single-gain PID baseline maps a single error signal (power deviation) to a rod displacement through a fixed proportional gain, without reasoning about alternative actuation strategies or switching between control banks based on context. When the primary bank (Bank 2) saturates or when large power changes exceed its reactivity range, this one-shot mapping has no mechanism to discover or deploy alternative solutions. The agentic model, by contrast, learns the full manifold of admissible strategies and selects among them based on task requirements.

The Direct LoRA baseline's catastrophic failure (mean error exceeding 100\%) empirically validates our curriculum design. Without Phase 1 grammar learning, the model has no structural prior to guide its hypothesis space. It attempts to learn both the syntax of valid control commands and the mapping from power targets to actions simultaneously, resulting in complete failure. This confirms that the two-phase separation is not merely a computational convenience but a fundamental architectural requirement for this domain.

Figure~\ref{fig:comparison}b stratifies success rates by power-change magnitude (small, medium, large). The single-gain PID baseline exhibits a clear inverse relationship between maneuver magnitude and reliability. While it performs adequately in the small regime (where incremental adjustments keep Bank 2 within its linear operating range), its success rate progressively declines in the medium regime and suffers severe degradation in the large regime. This monotonic drop illustrates the limitations of a single-shot linear mapping when facing the actuator constraints and non-linear kinetics inherent to large reactivity insertions. 

Formally, our single-gain baseline computes the rod-displacement command as
\begin{equation}
\Delta s = K_p\,(P_{\text{target}} - P_{\text{initial}}),
\end{equation}
where $K_p$ is fitted once on a 6-shot symmetric calibration set. This one-shot formulation, dictated by the simulator's batch-transient interface, assumes a locally linear, monotonic dependence of terminal power on net rod displacement and unlimited actuation authority within the rod range.

Conversely, the proposed model maintains consistently high reliability across all regimes, effectively decoupling control performance from the magnitude of the requested power shift. This regime-invariant performance is consistent with the behavior expected of a domain-specific foundation model. The model has learned not a single control law but a portfolio of strategies, each suited to different operational contexts. Small shifts deploy fine adjustments through Bank 2. Large shifts engage Bank 1 or coordinate both banks. The model's internal representation captures this contextual structure and selects appropriately at runtime.

The log-scale error distribution in Figure~\ref{fig:comparison}c reveals a qualitative distinction in control precision. While the single-gain PID baseline achieves a respectable median error of 7.08\%, its distribution is heavy-tailed with a maximum error of 25.59\%. This indicates that while the baseline works on average, it is prone to catastrophic excursions. The 95th percentile error is approximately 22\%, meaning that 1 in 20 operations could result in significant power deviation requiring operator intervention or protective action.

The proposed 100K model demonstrates extreme precision with a median error of 0.31\% and a mean of 0.61\%, resulting in a distribution tightly clustered near zero. This precision is a direct consequence of the high-reliability action region under the sampled distribution and tail-risk collapse described earlier. The model has learned to avoid brittle regions of the action space and concentrate on robust, high-success strategies. The distribution's sharp peak at zero reflects policy stabilization: the model consistently proposes actions that land within the solution manifold $\mathcal{M}_{\text{sol}}(o^*)$.

From a safety perspective, the CDF in Figure~\ref{fig:comparison}d is the most critical metric. The single-gain PID baseline's CDF rises gradually, retaining an error of approximately 22\% even at the 95th percentile. This gradual rise indicates persistent tail risk. In contrast, the proposed model's CDF exhibits a step function-like rise very close to 0\%, with the 99th percentile error remaining as low as 7.22\%.

Define the tail-risk reduction ratio as:
\begin{equation}
\mathcal{R}_{95} = \frac{Q_{95}(\text{PID})}{Q_{95}(\text{Proposed})} \approx \frac{22\%}{1.3\%} = 16.
\end{equation}
The proposed model achieves a 16-fold reduction in 95th percentile error relative to the single-gain PID baseline. This is not incremental improvement; it is structural elimination of tail risk through learned high-reliability action region under the sampled distribution
 concentration.

To verify that the gap between the learned policy and classical control is not an artifact of the single-gain baseline, we additionally implemented a gain-scheduled PID controller with six regime-dependent proportional gains (small/medium/large $\times$ increase/decrease), tuned via a 40-shot calibration sweep on a separate calibration set and evaluated on the same 2{,}000-run closed-loop protocol. The gain-scheduled controller achieves $S(\pm 5\%) = 75.8\%$, a 35.1-percentage-point improvement over the single-gain baseline (40.7\%) while remaining 21.6 percentage points below the foundation model (97.4\%). Five of six regimes reach $\geq 66\%$ at $\pm 5\%$ (four at 100\%); the residual gap to the foundation model is concentrated in the large-decrease regime (16.9\%), where the reactor's reactivity worth saturates at deep rod insertion and no proportional control law can recover the lost authority. This localizes the foundation model's advantage to a specific physical mechanism rather than to weakness of the baseline.

Beyond accuracy and safety, the proposed approach offers substantial computational advantages over optimization-based classical control. Model-predictive control (MPC), the state-of-the-art for constrained optimization in nuclear systems, solves a constrained optimization problem at each timestep:
\begin{equation}
\min_{\mathbf{u}} \sum_{k=0}^{N} \left[ \|\mathbf{x}(k) - \mathbf{x}^*\|_Q^2 + \|\mathbf{u}(k)\|_R^2 \right]
\end{equation}
subject to plant dynamics, actuator constraints, and safety bounds. For high-fidelity reactor models, this optimization requires minutes to hours per query, especially when exploring multiple alternative maneuvers for what-if analysis.

The proposed model achieves training in 6 hours on a single NVIDIA RTX 3070 GPU (8GB VRAM) for the 100K model, inference latency in milliseconds per control proposal, and generation of dozens of candidate strategies in seconds for rapid screening in simulator-in-the-loop workflows. This order-of-magnitude improvement in operational tempo enables a fundamentally different workflow.Rather than running expensive optimizations sequentially, operators can generate a diverse portfolio of candidate strategies, validate them in parallel using the digital twin, and select the best option based on operational constraints, equipment status, and procedural requirements. The model becomes a proposal generator that amplifies human decision-making rather than replacing it.

The Direct LoRA baseline failure (mean error $>100\%$) validates the necessity of the two-phase curriculum. The single-gain proportional baseline (40.7\% success) and its gain-scheduled variant (75.8\%) confirm that simple feedback control is insufficient on this class of problem, while also showing that the residual gap to the foundation model is robust to baseline strength: even a regime-aware proportional controller plateaus at 75.8\% because of reactor reactivity saturation in the large-decrease regime. Within the scope of this study, the proposed agentic AI learns a low-variance control policy through offline data scaling and structural pretraining, achieving an approximately 16-fold tail-error reduction relative to the single-gain baseline and 97.4\% success across power-change regimes on the fixed-initialization, single-step task evaluated here.

\subsection*{Model actuation patterns at runtime}
\label{sec:actuation_pattern}

The transition from labeled examples to autonomous actuation is a defining test of whether a model behaves as an agent rather than a static function approximator. In a domain-specific foundation model, we expect the trained system not simply to reproduce the empirical distribution of actions in the dataset, but to discover and deploy stable behavioral strategies that reflect underlying physical structure, operational efficiencies, or control-theoretic regularities. The degree to which the model deviates from the training distribution while maintaining high success is therefore a direct measure of its agenticity.

\begin{figure}[htbp]
  \centering
  \includegraphics[width=\linewidth]{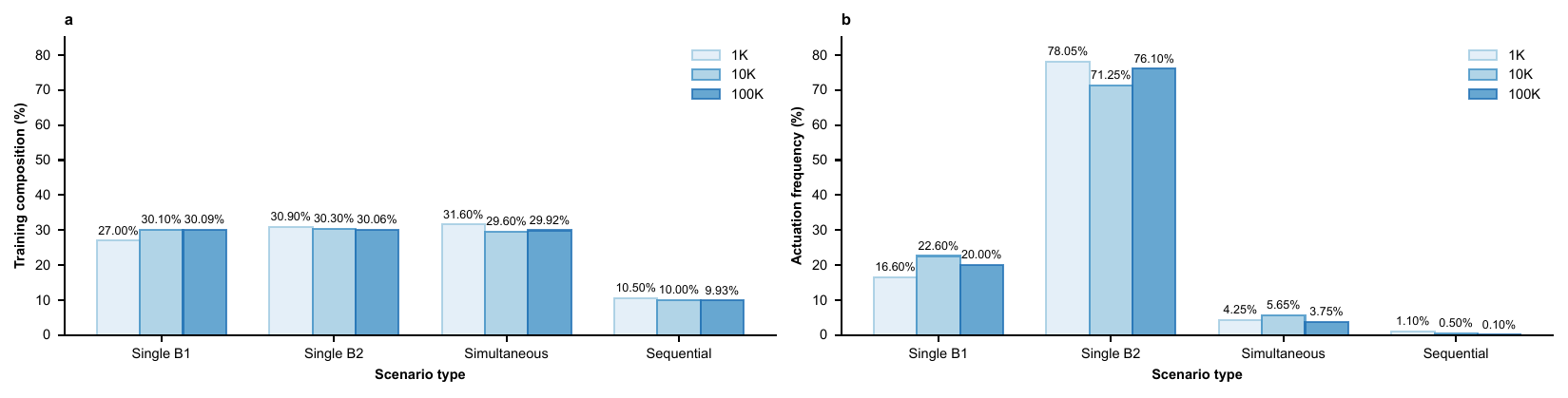}
\caption{\textbf{Runtime actuation patterns and success rates across model scales}. \textbf{a.} The proportional distribution of actuation patterns in the 1K, 10K, and 100K training datasets, engineered to be balanced. \textbf{b.} The distribution of actuation patterns executed during 2,000 simulator-run evaluations for each model scale. The pronounced divergence between (a) and (b), particularly the twofold enrichment of single-bank strategies and the strong concentration on single\_b2, reflects emergent policy formation. The 100K model deviates most sharply from the training distribution, indicating that agentic optimization becomes stronger as model capacity and data scale increase.}
  \label{fig:komodo-actuation}
\end{figure}

Figure~\ref{fig:komodo-actuation} highlights this contrast by comparing the composition of the training corpora (Figure~\ref{fig:komodo-actuation}a) with the model's runtime actuation patterns during 2,000 fully closed-loop validation runs (Figure~\ref{fig:komodo-actuation}b).

Four actuation patterns were distinguished based on which control banks were moved: \texttt{single\_b1} (only bank 1 actuated), \texttt{single\_b2} (only bank 2 actuated), \texttt{simultaneous} (both banks initiated at the same time), and \texttt{sequential} (both banks moved but with different start times). This taxonomy is a methodological choice of the present study, derived from elementary PWR rod-actuation conventions in which coarse and fine banks can be moved independently or in coordination~\cite{johnson2010modeling}; it is not a mapping of plant-specific operating procedures, which differ across reactor designs and licensing regimes. The definition of an \emph{admissible} strategy in this paper is operational rather than procedural: within each taxonomic class, any six-parameter command that (i) lies within actuator limits (position $\in [0,180]$ steps, $v > 0$, $t \geq 0$) and (ii) drives the simulator to the target power within tolerance is treated as admissible. Plant-specific procedural constraints (technical specifications, surveillance requirements, operator action limits) would constitute an additional filter in any deployment context and are intentionally outside the present training corpus; this limitation is discussed explicitly in the Discussion (Limitations). The training datasets were explicitly engineered to be balanced: approximately 60\% single-bank maneuvers and 40\% two-bank maneuvers, with each of the single-bank cases (\texttt{single\_b1} and \texttt{single\_b2}) comprising roughly 30\%, and the two-bank cases (\texttt{simultaneous}, \texttt{sequential}) comprising about 30 and about 10\%, respectively. The 1K, 10K, and 100K corpora closely matched this target composition (Figure~\ref{fig:komodo-actuation}a), ensuring that no dominant actuation pattern could trivially bias the model.

To distinguish agentic policy formation from passive imitation, we compare the runtime distribution against the balanced 30/30/30/10 training mixture:
\begin{equation}
H_0: D_{\text{KL}}(P_{\text{runtime}}\,\|\,P_{\text{train}}) \approx 0 \quad (\text{frequency-matching imitator}), \quad
H_1: D_{\text{KL}}(P_{\text{runtime}}\,\|\,P_{\text{train}}) \gg 0 \quad
\end{equation}
The runtime distributions yield $D_{\text{KL}} = 0.54$, $0.44$, and $0.55$~nats at 1K, 10K, and 100K (Table~\ref{tab:entropy_kl_recomputed}). At every scale $D_{\text{KL}}$ is substantially above zero, supporting $H_1$; the bulk of probability is assigned to \texttt{single\_b2} (76.14\% at 100K) rather than to the training-frequency mode, inconsistent with passive imitation. The 100K model concentrates 76.1\% of runtime execution on single\_b2 despite only 30\% training exposure, constituting formal evidence of optimization: the model discovered and exploited a high-success strategy (99.5\% reliability, Figure~\ref{fig:komodo-scenario}c) through learned structural preferences, not dataset frequency. The 100K model, with greatest capacity, uses that capacity to optimize away from the balanced training mixture toward a skewed, high-success strategy. This is a characteristic of agentic behavior: the model has learned that optimizing for task success implies concentrating on a subset of strategies, even at the cost of deviating from the data it was trained on.

In stark contrast to the balanced training composition, the model's runtime behavior demonstrates a striking and highly consistent departure from this design. Across all scales, the model exhibits a strong preference for single-bank solutions, particularly \texttt{single\_b2}. This preference is observed in 78.05\% of runs for the 1K model, 71.25\% for the 10K model, and 76.1\% for the 100K model. Even \texttt{single\_b1}, although used far less frequently, accounts for an additional 16.60 to 22.60\% of model decisions. Aggregated across all three models, single-bank strategies constitute 94.65, 93.85, and 96.1\% of all executed strategies for the 1K, 10K, and 100K models, respectively. This is more than a twofold enrichment relative to the 60\% single-bank composition of the training data and a significant deviation from the nearly uniform 30/30/30/10 training distribution.

This divergence characterizes emergent policy formation: the model actively selects actions based on their efficacy in achieving task objectives rather than passively sampling from the training distribution. The model has discovered that the space of high-success control strategies is much smaller and more structured than the space of feasible strategies, and concentrates its probability mass accordingly.

The model learns that initiating motion in the already-partially-inserted bank (bank 2 at 100 steps) is the most efficient and reliable mechanism for producing a wide range of power changes across nominal PWR operating conditions. Because bank 1 begins fully withdrawn (180 steps), engaging it often requires a larger traversal of the rod range to produce equivalent reactivity effects. The model has internalized this asymmetry through data-driven inference and distilled it into a robust control strategy: act primarily through bank 2, resorting to bank 1 only when the required magnitude or direction of power change exceeds the efficient actuation range of bank 2. This emerges from the structure of the data and the nonlinear dynamics of the reactor system rather than from explicit programming.

This convergence onto a physically meaningful strategy is notable for two reasons. First, the preference for \texttt{single\_b2} is not present in the training distribution; the training data explicitly provided equal representation of single-bank and two-bank scenarios. The emergence of a dominant actuation class indicates that the model has developed a representation of the control manifold that reflects physical structure rather than dataset frequency. The 1K runtime distribution is concentrated on \texttt{single\_b2} ($\approx 78\%$). We computed $H(P) = -\sum_{a \in \mathcal{A}} P(a) \log P(a)$ directly from the raw runtime histograms in Figure~\ref{fig:komodo-actuation}b (with the convention $0\log 0 \equiv 0$); the values are reported in Table~\ref{tab:entropy_kl_recomputed} and in the caption of Figure~\ref{fig:komodo-actuation}.

All three runtime distributions are far from uniform (the model concentrates on single-bank strategies at every scale), and the marginal entropy over actuation classes does not vary monotonically with scale. We characterize the scaling effect at the more informative level of \emph{conditional} reliability per actuation class (Figure~\ref{fig:komodo-scenario}): the share of attempted strategies that succeed rises sharply with scale even when the marginal distribution over strategies does not. The model is not a passive memorizer of the balanced 30/30/30/10 training mixture (see the KL analysis in E2 below).

\begin{table}[h]
\centering
\caption{Computed policy entropy and KL divergence from runtime distributions in Figure~\ref{fig:komodo-actuation}b. All values computed directly from the raw histograms with the convention $0\log 0 \equiv 0$; empty bins use $\epsilon = 10^{-3}$ smoothing where required by KL. The analysis script is released alongside the manuscript code.}
\label{tab:entropy_kl_recomputed}
\begin{tabular}{lccc}
\toprule
Scale & $H(P_{\text{runtime}})$ [nats] & $H_{\max} = \ln 4$ [nats] & $D_{\text{KL}}(P_{\text{runtime}}\,\|\,P_{\text{train}})$ [nats] \\
\midrule
1K   & 0.6754 & 1.386 & 0.5407 \\
10K  & 0.7665 & 1.386 & 0.4430 \\
100K & 0.6596 & 1.386 & 0.5455 \\
\bottomrule
\end{tabular}
\end{table}

Second, the strategy recapitulates real operator practice. In full-scale PWRs, operators preferentially actuate the most effective rod bank, typically the one already positioned in the region of steepest reactivity slope, before engaging multiple banks in coupled motion. This alignment between learned AI policy and human operator heuristics reflects the fact that both optimize over the same underlying physical constraints. Through data-driven learning, the model has rediscovered principles that human reactor operators learned through decades of operational experience.

The model enacts consistent, outcome-oriented behaviors that reflect an underlying understanding of physical dynamics. Unlike a static mapping, which always produces the same distribution of outputs regardless of effectiveness, an agentic model identifies and repeatedly applies a stable policy that minimizes risk and maximizes control accuracy. This transition from reproducing distributions (memorization) to deploying a preferred actuation strategy (optimization) characterizes foundation models as they scale: given sufficient data diversity, they discover low-dimensional, high-utility structures embedded within high-dimensional action spaces.

Crucially, this preference for single-bank control is not detrimental to performance. On the contrary, the 100K model, despite using \texttt{single\_b2} in 76.1\% of all runs, achieves perfect success in the medium and large power-change bins and 86\% success in the most challenging small-change bin. This demonstrates that the model's policy specialization is not a collapse of capability; it is a refinement of control strategy. The model has discovered that the high-dimensional space of feasible strategies can be substantially compressed without sacrificing performance. In fact, the compression improves robustness: simpler strategies have fewer degrees of freedom, fewer sources of error, and more consistent outcomes.

The actuation results substantiate two fundamental claims. The model behaves as an agent learning a policy, not as a conditional density estimator reproducing training frequencies—agents optimize, density estimators approximate. The model exhibits structural simplification and regime specialization characteristic of domain-specific foundation models, organizing the high-dimensional control space into a coherent, reproducible, and physically grounded action manifold. Foundation models compress and regularize their internal representation of the domain as they scale rather than simply expanding. SmolLM2-360M discovers emergent rules for efficient reactor control rather than merely imitating labeled examples, demonstrating stable, low-variance actuation strategies aligned with the physics and operational heuristics of real nuclear systems.

\subsection*{Failure analysis}

\begin{figure}[htbp]
  \centering
  \includegraphics[width=\linewidth]{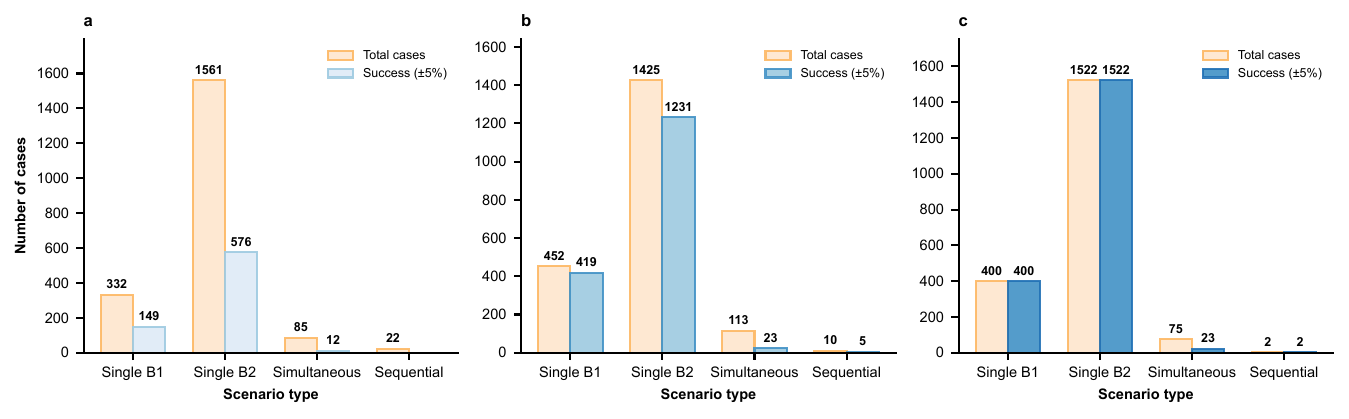}
\caption{\textbf{Validation case count and success rate by actuation pattern and model scale.} Each panel shows the total number of runtime attempts and the number of successful cases ($\pm$5\% tolerance) for the four actuation patterns (\texttt{single\_b1}, \texttt{single\_b2}, \texttt{simultaneous}, \texttt{sequential}) at one model scale. \textbf{a.} 1K-scenario model achieves success rates below 50\% across all four actuation classes, with no reliable strategy emerging; the learned policy is unstructured and unstable. \textbf{b.} 10K-scenario model achieves strong performance on \texttt{single\_b1} (419/452) and \texttt{single\_b2} (1{,}231/1{,}425) but succeeds far less often on \texttt{simultaneous} (23/113) and \texttt{sequential} (5/10), indicating that the model has discovered single-bank actuation to be safer and more effective. \textbf{c.} 100K-scenario model attains perfect reliability on \texttt{single\_b1}, \texttt{single\_b2}, and \texttt{sequential} cases; only \texttt{simultaneous} remains imperfect (23 successes out of 75 attempts), reflecting that the model deploys two-bank coordination sparingly as a specialized tool. The evolution from uniform low success (1K) to selective excellence (10K) to near-perfect discrimination (100K) demonstrates that the model learns a hierarchical risk structure. It identifies safe regions and preferentially operates within them, deploying higher-risk strategies only when necessary.}
  \label{fig:komodo-scenario}
\end{figure}

\begin{figure}[htbp]
  \centering
  \includegraphics[width=\linewidth]{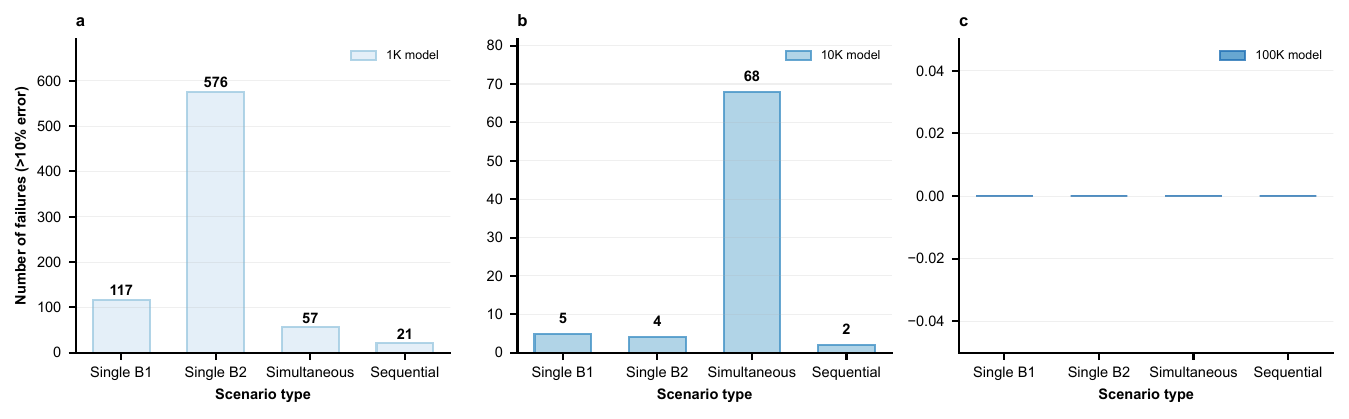}
  \caption{\textbf{Distribution of severe failures (greater than 10\% error) across actuation patterns and dataset scales.} Each panel shows the count of severe failures (terminal-power deviation $>\pm 10\%$ from target) per actuation pattern at one model scale. \textbf{a.} 1K-scenario model produces 771 severe failures, with 576 concentrated in \texttt{single\_b2}; failures are widespread across all four actuation classes, indicating that the model lacks any learned distinction between safe and unsafe actions. \textbf{b.} 10K-scenario model reduces severe failures 9.7-fold to 89 events, with 68 now concentrated in \texttt{simultaneous} actuation; the model has learned that single-bank strategies are safe but has not yet mastered two-bank coordination. \textbf{c.} 100K-scenario model eliminates all severe failures, demonstrating complete internalization of the high-reliability action region within the sampled distribution. Scaling from 1K to 100K induces a collapse of catastrophic outliers, a prerequisite for agent-level reliability in safety-critical domains.}
  \label{fig:komodo-failure}
\end{figure}

Failure analysis reveals how the model organizes its action space and whether its behavior reflects purposeful policy formation or stochastic imitation. The structure, concentration, and regime-dependence of failures indicate whether the model learns a high-reliability action region under the sampled distribution low-dimensional subspace of the full action space within which high-success control strategies are concentrated. Figures~\ref{fig:komodo-scenario} and~\ref{fig:komodo-failure} characterize failure modes across scales.

Figure~\ref{fig:komodo-scenario} contrasts, for each actuation pattern, the number of times a model attempted a given scenario versus the number of times it succeeded within the $\pm$5\% tolerance. Although all three models attempt patterns in broadly similar proportions, reflecting their shared preference for single-bank solutions as shown earlier in Figure~\ref{fig:komodo-actuation}, they differ sharply in the reliability with which they execute these strategies.

The 1K model (Figure~\ref{fig:komodo-scenario}a) achieves success rates below 50\% across all four actuation classes. No reliable strategy emerges; all patterns fail with comparable probability. The learned policy is unstructured, undifferentiated, and unstable. The model has not learned even the coarse topology of the control action space.

The 10K model (Figure~\ref{fig:komodo-scenario}b) marks a decisive shift. It achieves strong performance on \texttt{single\_b1} (419/452) and \texttt{single\_b2} (1,231/1,425), but succeeds far less often on \texttt{simultaneous} (23/113) and \texttt{sequential} (5/10) patterns. The model selectively excels in low-dimensional, robustly controllable regions of the action space and struggles in high-dimensional, delicate regions requiring precise coordination of two banks. The model has discovered that single-bank actuation is consistently safer and more effective, and gravitates toward these strategies at runtime. The persistent failures in the \texttt{simultaneous} case reflect its sensitivity to small timing or speed errors, an inherently fragile actuation pattern that offers little margin for correction in a nonlinear reactor environment. The model is learning to avoid this fragile region of the control space.

The 100K model (Figure~\ref{fig:komodo-scenario}c) completes this transition from pattern reproduction to agentic decision-making. For \texttt{single\_b1}, \texttt{single\_b2}, and even \texttt{sequential} cases, the success bars match the total bars exactly, indicating perfect reliability across 2,000 runs. Only the \texttt{simultaneous} case remains imperfect, with 23 successes out of 75 attempts. This selective vulnerability is revealing: the model does not avoid \texttt{simultaneous} actions entirely, but treats them as a specialized tool deployed sparingly. The model has learned that most power demands can be met via simple, robust single-bank or sequential strategies, but in rare cases, approximately 3.75\% of tasks (75 out of 2000), the model recognizes that simultaneous two-bank action is necessary and attempts it. 

This is analogous to how experienced human operators behave: they primarily rely on the most controllable bank and invoke more complex, coupled motion only when absolutely necessary. This alignment between emergent AI policy and human operational heuristics further supports the interpretation of the 100K model as a physics-aware agent rather than a memorization-based regressor. It is not simply replicating training data; it is applying learned principles to make strategic decisions about risk and efficacy.

To evaluate safety-critical behavior more directly, we examined the distribution of severe failures, defined as runs whose terminal power deviated by more than $\pm$10\% from the target. Figure~\ref{fig:komodo-failure} summarizes these events. Severe failures represent catastrophic errors that would trigger protective actions in a real reactor. Their presence and distribution are diagnostic of whether the model has internalized a notion of safe versus unsafe control strategies. The differences across scales are stark.

The 1K model (Figure~\ref{fig:komodo-failure}a) produces 771 severe failures, reflecting deep structural instability. Most occur in \texttt{single\_b2} (576 failures), the simplest maneuver. This is counterintuitive: one would expect simple maneuvers to be more reliable, not less. The fact that they fail most often indicates that the model has learned no meaningful structure; it cannot even reliably execute the simplest strategies. This suggests that the 1K model lacks any internal distinction between safe and unsafe actions. From the model's learned perspective, the control action space is essentially a featureless landscape where all strategies are equally unreliable. Such behavior is incompatible with any safety-critical application.

The 10K model (Figure~\ref{fig:komodo-failure}b) demonstrates a dramatic reduction in catastrophic events, falling to 89 total severe failures. This is a 9.7-fold reduction, a massive improvement. More importantly, the failures are no longer uniformly distributed. Instead, they concentrate overwhelmingly within a single scenario: \texttt{simultaneous} actuation (68 failures). This concentration is highly diagnostic of policy maturation. The model has learned to navigate safe regions of the control space, reducing severe failures in \texttt{single\_b2} from 576 to only 4, but has not yet mastered the brittle, coordinated dynamics of two-bank motion. 

This segregation of risk indicates high-reliability action region under the sampled distribution formation: the model has begun to partition the action space into reliable and unreliable regions and adjust its behavior accordingly. It has learned that \texttt{single\_b2} is safe and concentrates its efforts there; it has learned that \texttt{simultaneous} is risky and avoids it when possible. Through exposure to diverse control scenarios and simulator feedback, the model has internalized a map of the safe and unsafe regions of the reactor control landscape.

The 100K model (Figure~\ref{fig:komodo-failure}c) eliminates all severe failures entirely. This complete elimination of catastrophic tail risk is not an incremental improvement; it is structural transformation. This collapse of catastrophic tail risk is a defining requirement for a reliable Physical AI system.
 The disappearance of outliers indicates that the model has not only learned which actions work, but also learned to avoid regions of the control space where small perturbations can lead to runaway errors. The 100K model has internalized a precise high-reliability action region under the sampled distribution, a low-dimensional subspace of the high-dimensional action space within which all actions are reliable. It does not merely avoid \texttt{simultaneous} actuation; it has learned the precise conditions under which each action class succeeds or fails, and it has organized its internal representations to concentrate probability mass on regions of guaranteed success. This is analogous to the high-reliability action region under the sampled distribution formation observed in high-scale foundation models for robotics and climate, where large, diverse datasets enable the model to internalize physically meaningful invariances that eliminate pathological behaviors.

To formalize the progression toward reliability emerging from scale, we define the normalized failure concentration as:
\begin{equation}
\mathcal{C} =
\frac{\max(\text{failures by actuation class})}{\text{total failures}}.
\end{equation}

This metric quantifies how concentrated failures are. A highly structured model concentrates failures in a few brittle regimes ($\mathcal{C}$ close to 1); an unstructured model distributes failures uniformly ($\mathcal{C}$ close to 0.25 for 4 classes). From the data:
\begin{itemize}
    \item 1K model: $\mathcal{C} = 576/771 = 0.747$ (failures concentrated in \texttt{single\_b2}, but failures still numerous overall)
    \item 10K model: $\mathcal{C} = 68/89 = 0.764$ (failures highly concentrated in \texttt{simultaneous})
    \item 100K model: $\mathcal{C} = 0/0$ undefined (no failures exist to concentrate)
\end{itemize}

The increasing $\mathcal{C}$ through the 10K model indicates growing structural organization: failures become predictable, concentrated, and avoidable. The elimination of failures entirely in the 100K model indicates that the high-reliability action region under the sampled distribution has been fully internalized. The model has learned to operate exclusively within the safe region.

Formally, denote the true success probability of action $a$ as $\mathbb{P}(s \mid a)$. The model learns an estimate $\hat{\mathbb{P}}_n(s \mid a)$ from $n$ training samples. At low scales, this estimate is noisy and unreliable everywhere. At high scales, it becomes accurate and sharply distinguishes high-success from low-success regions. The runtime policy $p_\theta(a \mid \text{target})$ implicitly follows this learned success probability. As the estimate improves with scale, the policy naturally concentrates on high-success actions, and severe failures, which occur in low-success regions, automatically become rare.

This mechanism is distinct from explicit reward shaping or constraint enforcement. It is purely a consequence of the model learning an accurate map of which actions succeed and which fail, then concentrating its probability mass on successful regions. Reliability emerges not from punishment but from learned structure.

The progression in failure modes from 1K to 100K reveals the structural evolution of the model from an unstable mapping to a physically aligned agent. The 1K model behaves as an unstructured approximator with failures uniformly distributed across all actuation classes. The 10K model identifies safe regions (single-bank) and concentrates catastrophic errors in complex regimes (\texttt{simultaneous}), beginning to partition the action space. The 100K model eliminates catastrophic errors entirely and deploys multi-bank strategies sparingly and strategically, learning not merely what to do but what not to do.

With sufficient scale, the model reorganizes its action space, aligns its strategy with the physics of reactor operation, and avoids pathological behaviors autonomously. The 100K model has internalized a latent high-reliability action region under the sampled distribution and operates on it consistently, rediscovering through data-driven learning the principles of safe reactor control that took human operators decades to learn.

However, this concentration on a high-reliability action region under the sampled distribution is currently context-specific: it applies to single-step power maneuvers under the specific initialization pattern (Bank~1 at 180, Bank~2 at 100) used in the main training set. Our mixed-initialization analysis shows that the same network conditioned on a different initialization shifts its concentration to the mirrored bank. A foundation model with broader operational coverage would additionally demonstrate transfer across diverse initial conditions, actuation configurations, and reactor types, none of which is established here. The results here establish proof of concept: offline data scaling can drive the emergence of stable, physics-aware policies without online exploration or explicit safety rewards (addressing the exploration-risk barrier from the Introduction). Extending this capability to the full scope of nuclear operations (multi-step procedures, mixed actuators such as rods plus boron plus feedwater, diverse reactor designs) defines the path toward a complete domain-specific foundation model. We have demonstrated the mechanism (scaling plus curriculum plus Physical AI validation); the generalization breadth remains to be proven.

In short, the failure analysis shows that the model learns not just what to do, but what not to do, an essential criterion for Agentic Physical AI and a foundational requirement for safe control in nuclear systems.

\subsection*{Generalization and Architectural Flexibility Assessment}

\begin{figure*}[t]
  \centering
  \includegraphics[width=\linewidth]{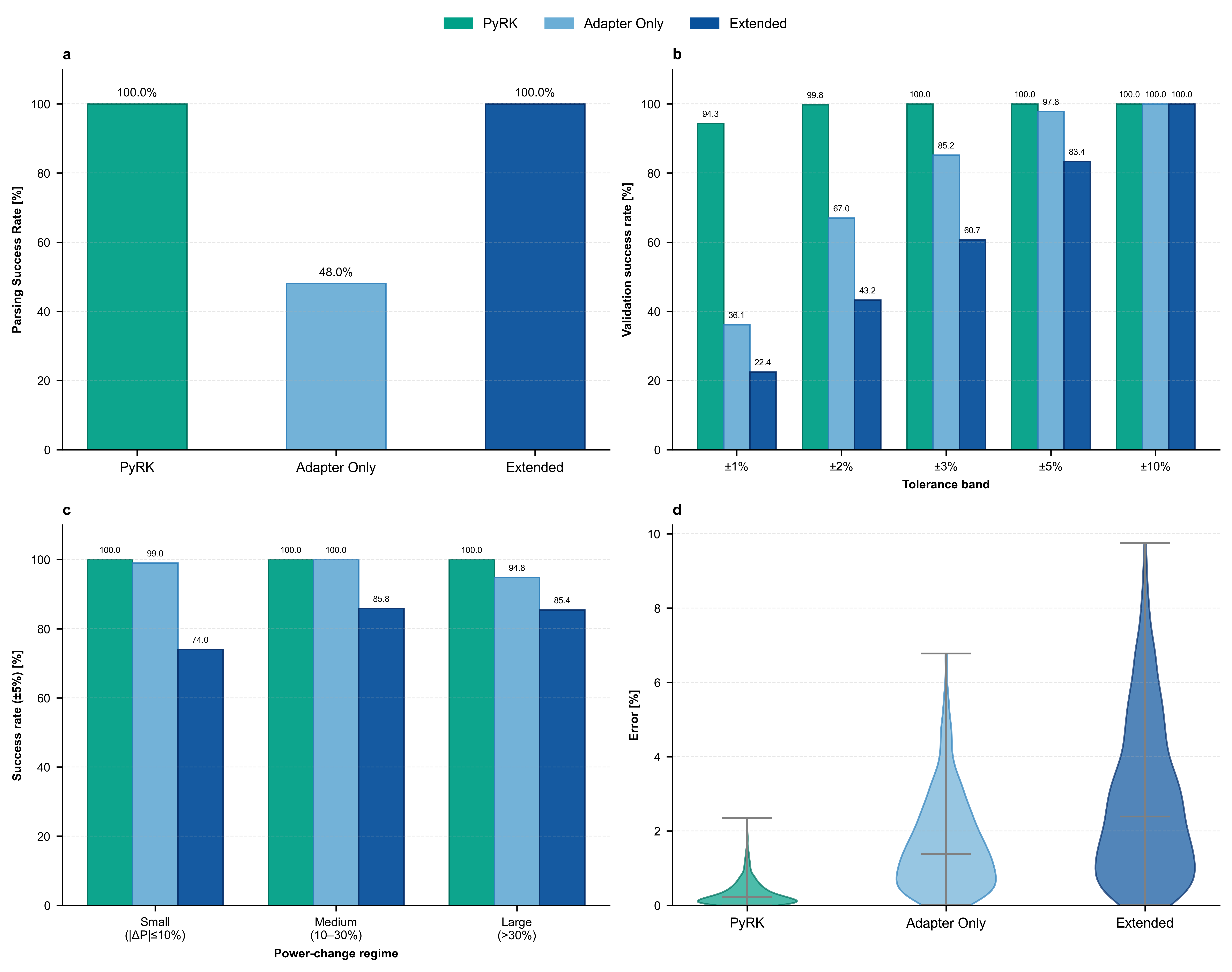}
  \caption{\textbf{Performance comparison of generalization (PyRK) and architectural extension (Variable Window) models.}
  \textbf{a.} Parsing success rates reveal that the Adapter Only approach fails to maintain valid syntax (48\%), whereas PyRK and Extended models achieve 100\%.
  \textbf{b.} Validation success rates across tolerance bands show PyRK's superior precision, while the Extended model trades strict accuracy for flexible time-window handling.
  \textbf{c.} Regime-stratified success rates ($\pm$5\%) demonstrate robustness in PyRK, while the Extended model shows sensitivity in small and large maneuver regimes.
  \textbf{d.} Violin plots of error distributions highlight the minimal variance of the PyRK model compared to the broader uncertainty of the variable-window approaches.}
  \label{fig:extension_results}
\end{figure*}

To validate the scalability of the proposed framework beyond the specific constraints of the KOMODO simulator and fixed-time horizons, we evaluated the performance of the generalization (PyRK) and architectural extension (Variable Window) models. These experiments probe whether the Agentic Physical AI framework and the underlying toward-foundation-model design are specific to one simulator and fixed evaluation horizon, or whether they transfer to new physics and broader temporal structures. Figure~\ref{fig:extension_results} summarizes the comparative performance across parsing success, validation rates, and error distributions.

The PyRK model achieved 100\% parsing success and $>94\%$ validation success even at the strictest $\pm$1\% tolerance (Figure~\ref{fig:extension_results}b), maintaining near-perfect reliability across all power-change regimes (Figure~\ref{fig:extension_results}c) with minimal error variance (Figure~\ref{fig:extension_results}d). The Agentic Physical AI methodology transfers across distinct physical environments with minimal tuning. The same compact language model, trained under the two-phase curriculum, generalizes from KOMODO's spatial neutronics to PyRK's point kinetics with zero architectural modification.

The same two-phase curriculum (grammar learning plus task conditioning) that succeeded in KOMODO's spatial neutronics transfers to PyRK's point kinetics with zero architectural modification and minimal retraining. The model learned a reusable representation of inverse reactor kinetics that generalizes across physics formulations rather than memorizing KOMODO-specific patterns. This structural transfer distinguishes foundation models from task-specific controllers.

However, both KOMODO and PyRK are PWR-like systems with rod-based reactivity control; transfer to fundamentally different reactor types (liquid-salt with chemical shim, gas-cooled with variable coolant flow, fast reactors with sodium cooling) or mixed actuation modalities has not been demonstrated. The results demonstrate initial cross-simulator transfer within a reactor family. Extension to diverse reactor types and actuation modalities will establish universal domain coverage.

The architectural extension to variable monitoring windows ($t_{\text{window}} \in [60, 100]$ seconds) revealed critical insights into the integration of continuous inputs.

Syntactic stability: The Adapter Only approach, which froze the language model backbone, failed to generate valid control syntax in 52\% of cases (48.0\% parsing success, Figure~\ref{fig:extension_results}a). In contrast, the Extended model (Adapter plus LoRA) restored 100\% parsing success. This finding suggests that projecting continuous variables into the embedding space requires joint optimization of the language model's weights (via LoRA) to accommodate the new input modality. A simple linear projection is insufficient for maintaining structural coherence.

Simply projecting continuous values into the embedding space fails if the language model's internal attention heads are not updated to interpret these new soft tokens. The success of the Extended model confirms that LoRA provides sufficient plasticity to align the pre-trained discrete representations with new continuous modalities. This finding establishes a blueprint for future expansions: incorporating real-time sensor readings or complex boundary conditions will likely require joint optimization of input adapters and backbone parameters (via LoRA), rather than treating the language model as a frozen black box.

Precision trade-off: While the Extended model successfully learned the grammar of variable-time control, it exhibited higher variance in terminal error compared to the fixed-horizon baseline (Figure~\ref{fig:extension_results}d). The success rate at the strict $\pm$1\% band dropped to 22.4\%, although it remained robust at the operational $\pm$5\% band (83.4\%). This performance dip reflects the increased complexity of the task: the agent must now infer not only the rod dynamics but also the optimal actuation timing relative to a floating evaluation horizon.

Overall, these results establish that the proposed foundation-model-inspired architecture is sufficiently flexible to adapt to new physics and variable boundary conditions, provided that the adaptation strategy allows for sufficient parameter updates to align the pre-trained representations with new input modalities. The PyRK generalization and the Variable Window experiments together demonstrate that the Agentic Physical AI approach is not overfitted to a single reactor model or temporal configuration. Instead, it provides a pathway toward domain-specific foundation models for nuclear systems, where a single compact language model backbone can be adapted, via structured curricula and lightweight adapters, to a range of reactor dynamics and control tasks.

These extensions demonstrate that the architecture accommodates variable temporal goals and continuous data streams, prerequisites for real-world deployment where operational constraints are fluid.

\subsection*{Robustness to varied initial bank positions}
\label{sec:varied_init}

The main experiments fix the initial rod configuration at Bank~1~$=$~180 and Bank~2~$=$~100, placing Bank~2 near the steep portion of the reactivity curve. Under this configuration, single-Bank-2 maneuvers are intrinsically more effective, and the fixed initialization could on its own account for some of the model's runtime preference for \texttt{single\_b2}. To separate genuine policy learning from exploitation of this specific initialization, we trained a mixed-initialization variant of the model and validated it across two distinct initialization regimes.

We constructed a 100{,}000-scenario corpus combining (a) a stratified 50{,}000-sample subset of the original default-initialization data (Bank~1~$=$~180, Bank~2~$=$~100) and (b) 50{,}000 newly generated KOMODO simulations at a mirrored initialization (Bank~1~$=$~100, Bank~2~$=$~180), with the sampling logic mirrored so that the structural role of the steep-gradient bank is preserved. The actuation-family balance (60/30/10 single-bank/simultaneous/sequential) was preserved within each subset.

To make the initialization explicit to the model, the input schema was extended from eight to ten tokens: $[\texttt{init\_b1}, \texttt{init\_b2}, P_{\text{initial}}, P_{\text{target}}, \texttt{b1\_pos}, \texttt{b1\_time}, \texttt{b1\_speed}, \texttt{b2\_pos}, \texttt{b2\_time}, \texttt{b2\_speed}]$. Phase~1 (grammar via CPT) and Phase~2 (task conditioning via LoRA) followed the identical curriculum, LoRA configuration ($r=32$, $\alpha=64$), and hyperparameters used for the main 100K model.

The mixed-init model was evaluated on 2{,}000 closed-loop KOMODO runs at each of the two configurations (default and mirror), using the same target-power distribution as the main 2{,}000-case test set.

\begin{figure}[h]
  \centering
  \includegraphics[width=\linewidth]{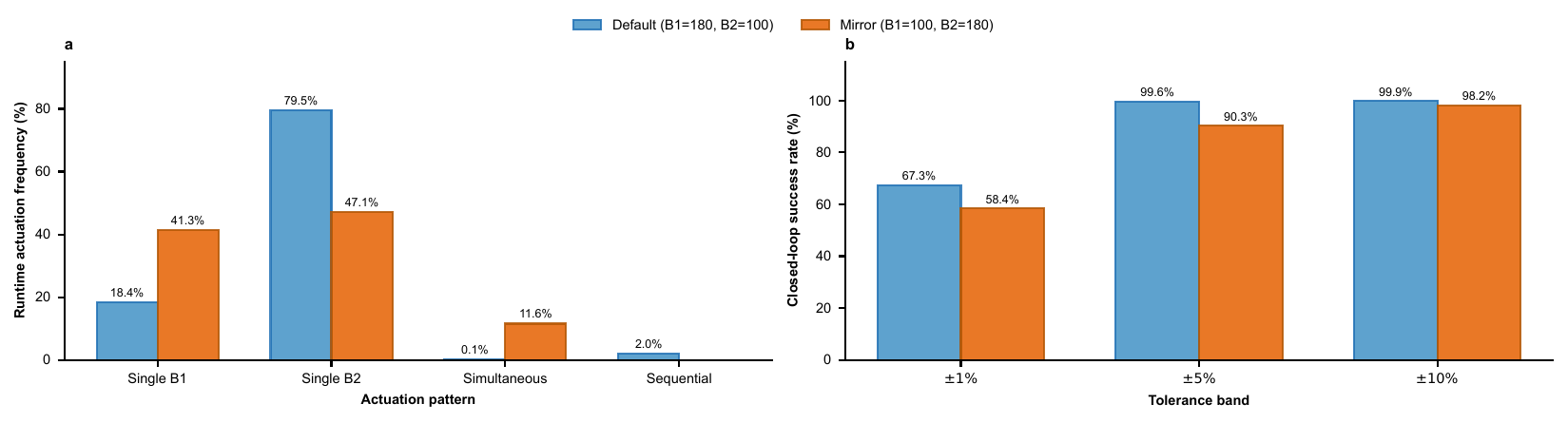}
  \caption{\textbf{Initial-rod-position variation experiment.}
  \textbf{a.} Runtime actuation distribution of the mixed-initialization model evaluated at the default initialization (Bank~1$=$180, Bank~2$=$100) and at the mirrored initialization (Bank~1$=$100, Bank~2$=$180). When the initialization is flipped, \texttt{single\_b1} usage more than doubles (18.4\%~$\to$~41.4\%) while \texttt{single\_b2} drops by 32.5 percentage points, and simultaneous coordination emerges as a compensation strategy (0.1\%~$\to$~11.6\%).
  \textbf{b.} Closed-loop success rate across tolerance bands. $\pm 5\%$ success is preserved above 90\% at both configurations (99.6\% default, 90.3\% mirror). The gap reflects an intrinsic bank-geometry asymmetry of the KOMODO reactor model rather than a model failure. $N = 2{,}000$ independent closed-loop runs per configuration.}
  \label{fig:varied_init_pattern}
\end{figure}

Figure~\ref{fig:varied_init_pattern}b and Table~\ref{tab:varied_init} report success rates. At the default initialization the mixed-init model achieves $S(\pm 5\%) = 99.6\%$ and $S(\pm 1\%) = 67.3\%$. At the mirrored initialization $S(\pm 5\%) = 90.3\%$ and $S(\pm 1\%) = 58.4\%$. The 9-percentage-point $\pm 5\%$ gap between configurations is attributable to a geometric asymmetry of the KOMODO reactor model, discussed below. The reduction in $\pm 1\%$ precision relative to the fixed-init regime is the cost of training across two distinct regimes.

This trade-off is the expected consequence of training a single model across two distinct initialization regimes rather than tuning to one, and is analogous to the gain-scheduled-versus-fixed-gain controller trade-off discussed earlier. The principal contribution of this section is the demonstrated context-dependence of the policy, not the strict-tolerance accuracy of the mixed-init variant.

The runtime actuation distribution differs substantially between the two configurations (Figure~\ref{fig:varied_init_pattern}a, Table~\ref{tab:varied_init_pattern}). At the default initialization the model is dominated by \texttt{single\_b2} (79.6\%, consistent with the 76.1\% reported earlier). At the mirrored initialization \texttt{single\_b1} usage more than doubles (18.4\%~$\to$~41.4\%) while \texttt{single\_b2} drops by 32.5 percentage points. Simultaneous two-bank coordination rises from 0.1\% to 11.6\%, an emergent compensation strategy that the model deploys when single-bank actuation is less reliable.

This shift is in the direction predicted by reactor physics. The model preferentially actuates whichever bank occupies the steep reactivity gradient region. The original \texttt{single\_b2} preference is therefore not an artifact of the fixed initialization but a context-dependent rule that targets the steeper-gradient bank, and the rule adapts when the initialization changes. Crucially, both runtime distributions are produced by the same network conditioned on the initial bank positions through its prompt (not by two separately trained models), so the shift reflects a single context-conditional decision rule learned within one network.

The 9-percentage-point gap between the two configurations, and the fact that the mirror is not perfectly symmetric (\texttt{single\_b1} reaches 41.4\% at mirror, smaller than the 79.6\% \texttt{single\_b2} share at default), reflect a known geometric asymmetry of the KOMODO reactor model. Inspecting the simulator's \texttt{\%CROD} card, each bank places two rod fingers in the 6$\times$6 quarter-core. Quarter-core reflective boundaries on the west and south edges then multiply corner fingers. Bank~1 places one of its fingers at the doubly-reflected south-west corner, yielding six rod assemblies in the full core versus four for Bank~2. The cross-section change is identical for both banks. The worth asymmetry is purely geometric. The mirror configuration therefore places the higher-worth bank in the active control role, increasing per-step reactivity sensitivity and making fine control inherently more demanding. The emergent reliance on simultaneous coordination at mirror (11.6\%) is the model's compensatory response to this physical asymmetry.

\begin{table}[h]
\centering
\caption{Mixed-initialization model: closed-loop success across two initialization regimes. All values from $N = 2{,}000$ independent closed-loop runs per regime.}
\label{tab:varied_init}
\begin{tabular}{lccccc}
\toprule
Initialization & $S(\pm 1\%)$ & $S(\pm 5\%)$ & $S(\pm 10\%)$ & mean err & max err \\
\midrule
Default (Bank~1$=$180, Bank~2$=$100) & 67.3\% & 99.6\% & 99.9\% & 0.95\% & 10.5\% \\
Mirror (Bank~1$=$100, Bank~2$=$180)  & 58.4\% & 90.3\% & 98.2\% & 2.04\% & 21.2\% \\
\midrule
Main 100K (default only, fixed-init prompt) & 92.0\% & 97.4\% & 100.0\% & 0.61\% & --- \\
\bottomrule
\end{tabular}
\end{table}

\begin{table}[h]
\centering
\caption{Mixed-initialization model: runtime actuation pattern shift across initialization regimes. Counts from $N = 2{,}000$ independent closed-loop runs per regime.}
\label{tab:varied_init_pattern}
\begin{tabular}{lcccc}
\toprule
Pattern & Default count & Default \% & Mirror count & Mirror \% \\
\midrule
\texttt{single\_b1}   & 367  & 18.4\% & 827 & 41.4\% \\
\texttt{single\_b2}   & 1591 & 79.6\% & 942 & 47.1\% \\
\texttt{simultaneous} & 2    & 0.1\%  & 231 & 11.6\% \\
\texttt{sequential}   & 40   & 2.0\%  & 0   & 0.0\%  \\
\bottomrule
\end{tabular}
\end{table}

\section*{Discussion}

A compact 360M-parameter language model functions as an Agentic Physical AI for nuclear reactor power control by discovering and deploying a low-variance, physics-aligned strategy through data scaling and outcome-centric validation. Agentic behavior manifests as systematic divergence between the balanced actuation mixture in training corpora (Figure~\ref{fig:komodo-actuation}a) and the sharply skewed runtime policy that preferentially selects single\_b2 (Figure~\ref{fig:komodo-actuation}b, 76.1\% usage). Scenario-specific reliability patterns (Figure~\ref{fig:komodo-scenario}) and severe-failure structure (Figure~\ref{fig:komodo-failure}) show that single\_b2 occupies a region of the control manifold that is both operationally efficient and measurably more robust under the simulator's initial-condition asymmetry.

This work establishes three interconnected claims. The model exhibits agentic behavior through policy optimization rather than memorization, operates as a Physical AI by learning outcome-driven inverse mappings rather than parameter-space regression, and displays early characteristics of a domain-specific foundation model through scaling-driven emergence, cross-simulator transfer, and structural compression. These properties form a mutually reinforcing system that addresses the three structural barriers to learning-enabled nuclear control identified in the Introduction: the many-to-one actuation problem, exploration risks in safety-critical domains, and the gap between numerical plausibility and execution-level correctness.

The central mechanistic insight is that these three properties constitute an integrated architectural loop rather than independent features.

First, agentic behavior enables Physical AI validation. Because the model learns a policy distribution over multiple admissible actions rather than predicting a single action vector, evaluation can be outcome-centric: success is defined by whether the reactor achieves the target power (Figure~\ref{fig:komodo-tolerance}, Figure~\ref{fig:komodo-combined}), not by whether predicted parameters match a specific labeled example (Table~\ref{tab:offline_mae_comparison}). This many-to-one mapping is what permits agentic selection. The model can concentrate probability mass on high-success strategies within the solution manifold $\mathcal{M}_{\text{sol}}(o^*)$ without penalty for deviating from any particular labeled trajectory.

Second, Physical AI validation drives agentic policy emergence. Outcome-centric evaluation creates a learning signal that rewards successful action selection regardless of parameter proximity to labels. This signal shapes the model to concentrate probability on high-reliability actions, the essence of policy formation. Critically, parameter-centric evaluation would penalize this divergence from labeled examples as variance; outcome-centric evaluation rewards it as exploration of the solution space. The 97.4\% closed-loop success (Figure~\ref{fig:komodo-tolerance}a) despite moderate offline MAE (Table~\ref{tab:offline_mae_comparison}) is direct evidence of this decoupling.

Third, scaling concentrates the runtime policy onto a sub-region of the action space within which closed-loop success is empirically high. At the 1K scale, the policy is noisy and exhibits 771 severe failures (Figure~\ref{fig:komodo-failure}a); at the 100K scale, severe failures on the sampled distribution are eliminated (Figure~\ref{fig:komodo-failure}c) and the runtime distribution concentrates on \texttt{single\_b2} (76.1\%, Figure~\ref{fig:komodo-actuation}b) with conditional reliability of 99.5\% for that strategy. We interpret this empirically: under outcome-centric training and a balanced training mixture, increasing data exposes the model to a wider sample of which actuations succeed in the simulator, and the learned policy assigns more probability mass to the strategies that do. We do not claim that this concentration is inevitable, nor that it should be interpreted through a thermodynamic analogy. It is the empirically observed scaling behavior of the present model on the present benchmark, and it is consistent with the broader pattern that foundation models compress training data into reusable structural priors.

In summary: agentic behavior becomes possible when evaluation is outcome-centric (Physical AI). Outcome-centric evaluation becomes trustworthy when grounded in deterministic physical simulation (addressing the execution-correctness gap). And the agentic policy becomes reliable and generalizable when the model is scaled on diverse offline data (foundation-model properties). These are not three separate contributions. They are three perspectives on a single coherent system for reliable, learning-enabled nuclear control.

Agentic behavior, as operationalized here, rests on three nested empirical claims that distinguish policy-forming agents from passive memorizers or reactive stimulus-response systems.

A frequency-matching imitator would produce a runtime distribution close to the balanced 30/30/30/10 training mixture, with $D_{\text{KL}} \approx 0$. The runtime distributions instead yield $D_{\text{KL}}(P_{\text{runtime}}\,\|\,P_{\text{train}}) = 0.54$, $0.44$, and $0.55$~nats at 1K, 10K, and 100K (Table~\ref{tab:entropy_kl_recomputed}). At every scale the runtime distribution differs materially from the training mixture, and the 100K model concentrates 76.14\% of runtime execution on \texttt{single\_b2} despite only 30\% training exposure (Figure~\ref{fig:komodo-actuation}b), inconsistent with passive imitation.

We computed $D_{\text{KL}}(P_{\text{runtime}}\,\|\,P_{\text{train}})$ directly from the raw runtime histograms (Table~\ref{tab:entropy_kl_recomputed}). At all three scales, the runtime distribution diverges substantially from the balanced 30/30/30/10 training mixture; in particular, the model assigns far less than 10\% probability to the \texttt{sequential} class that constitutes 10\% of the training corpus, and far more than 30\% probability to \texttt{single\_b2}. This is the qualitative pattern expected of an outcome-driven policy rather than of a frequency-matching imitator. The analysis script that produces Table~\ref{tab:entropy_kl_recomputed} is included with the source code.

The mechanism: The model implicitly learns to maximize success probability $\mathbb{P}(\text{success} \mid a)$, concentrating probability mass on actions that reliably achieve the physical objective. Because the training data is balanced (30/30/30/10 distribution across actuation families, Figure~\ref{fig:komodo-actuation}a), there is no dataset-level bias favoring single\_b2. The 76.1\% runtime concentration on single\_b2 (Figure~\ref{fig:komodo-actuation}b) emerges from the model discovering that this strategy occupies a robust region of the control manifold under the simulator's initial conditions (Bank 2 partially inserted at 100 steps, Bank 1 fully withdrawn at 180 steps). This is learned structural preference, not imitation. This constitutes policy optimization under an implicit reward function (closed-loop success), which is the defining feature of an agentic system.

Beyond simple policy concentration, the model exhibits strategic specialization. It deploys high-risk strategies (simultaneous two-bank) sparingly (3.75\% of attempts, Figure~\ref{fig:komodo-scenario}c), using them only when simpler strategies are insufficient.

Define the learned success rate for action class $a$ as:
\begin{equation}
\hat{\mathbb{P}}(s_a) = \frac{\text{number of successful runs using } a}{\text{number of runs attempting } a}.
\end{equation}

The 100K model's internal representation appears to encode an implicit risk hierarchy, as evidenced by Figure~\ref{fig:komodo-scenario}c:
\begin{equation}
\hat{\mathbb{P}}(s_{\text{single\_b2}}) \approx 0.995 \quad > \quad \hat{\mathbb{P}}(s_{\text{single\_b1}}) \approx 0.97 \quad > \quad \hat{\mathbb{P}}(s_{\text{simultaneous}}) \approx 0.31.
\end{equation}

Runtime decisions reflect this learned ordering: The model defaults to the safest action (single\_b2: 76.1\% usage), escalates to intermediate-risk actions (single\_b1: 20\%) when necessary, and deploys high-risk simultaneous coordination only in 3.75\% of cases where other strategies are predicted to fail (Figure~\ref{fig:komodo-actuation}b). This hierarchy is not programmed. It emerges from the model learning the conditional success structure of the action space through exposure to 100K diverse scenarios. This demonstrates that the model has internalized not merely which action is best on average, but the context-dependent reliability of each action class, a characteristic of strategic, risk-aware decision-making characteristic of intelligent agents.

A third marker of agentic behavior is what we term planning-by-preference: the model learns to prefer (assign high probability to) actions that lead to success without requiring explicit lookahead, value iteration, or tree search.

Formally, classical planning computes:
\begin{equation}
a^* = \arg\max_a \mathbb{E}[V(s') \mid s, a],
\end{equation}
requiring explicit simulation of future states $s'$. Planning-by-preference instead learns a direct policy mapping:
\begin{equation}
p_{\theta}(a \mid s, o^*) \approx p(a \mid \text{success at } o^*),
\end{equation}
where $o^*$ is the target outcome. The model has compressed the planning computation into a learned distribution that directly encodes which actions are likely to succeed for a given objective.

Evidence: The model generates control proposals in a single forward pass (low latency) yet achieves 97.4\% success (Figure~\ref{fig:komodo-tolerance}a), comparable to what iterative optimization methods achieve but without online search. The model has internalized the conditional distribution $p(a \mid \text{success})$ through offline training on success-labeled data. This is implicit value learning: the model knows which actions lead to high-value (successful) outcomes without explicit reward signals or rollouts.

Collectively, these three properties (policy optimization away from training distribution shown in Figure~\ref{fig:komodo-actuation}, strategic risk-aware specialization shown in Figure~\ref{fig:komodo-scenario}, and planning-by-preference) constitute a rigorous operational definition of Agentic AI that is empirically distinguishable from reactive stimulus-response and passive memorization.

Physical AI, as instantiated here, is grounded in two foundational principles that address the Introduction's third barrier (execution-correctness gap). This approach builds on emerging frameworks where physics-based validation serves as the ultimate arbiter of agent competence~\cite{kanwar2026phyplan}, extending agentic reasoning from manipulation tasks to safety-critical control domains.

Classical supervised learning in control optimizes parameter-space metrics:
\begin{equation}
\mathcal{L}_{\text{parameter}} = \mathbb{E}\left[\| a_{\text{label}} - a_{\text{predicted}} \|_2^2\right].
\end{equation}

Physical AI inverts this paradigm. Define the forward reactor dynamics as $\mathcal{F}: \mathcal{A} \to \mathcal{O}$, mapping actions to outcomes. Physical AI asks: which action(s) in $\mathcal{A}$ achieve the desired outcome $o^*$ when executed through $\mathcal{F}$? Formally:
\begin{equation}
\mathcal{L}_{\text{outcome}} = \mathbb{I}\left[\mathcal{F}(a_{\text{predicted}}) \notin \mathcal{B}_{\delta}(o^*)\right],
\end{equation}
where $\mathcal{B}_{\delta}(o^*)$ is the tolerance band around the target outcome (for example, $\pm$5\% power, Figure~\ref{fig:komodo-tolerance}).

This is fundamentally different. In many-to-one mappings (where multiple actions $\mathcal{A}^* = \{a : \mathcal{F}(a) = o^*\}$ yield the same outcome), parameter-centric metrics penalize all solutions except one arbitrary labeled example. Physical AI rewards all solutions that achieve the outcome, enabling the model to exploit the full solution manifold.

Empirical evidence of this distinction: The 100K model achieves 97.4\% closed-loop success (outcome metric, Figure~\ref{fig:komodo-tolerance}a) while maintaining moderate offline MAE of 26.95 steps for b1\_pos in single\_b1 cases (parameter metric, Table~\ref{tab:offline_mae_comparison}). Under parameter-centric logic, high MAE should predict poor performance. The divergence proves that the model has learned to reason about forward dynamics. 

Safety implication: A model optimized for parameter proximity might generate actions that match labeled examples numerically but produce dangerous outcomes when executed (the execution-correctness gap). A model validated through outcome-centric Physical AI naturally avoids parameter combinations that look similar to labels but would fail in the simulator. The 100K model's elimination of all severe failures (greater than 10\% error, Figure~\ref{fig:komodo-failure}c) is a direct consequence of this validation paradigm.

Physical AI requires that the model internalizes constraints imposed by underlying physics. In reactor control, these constraints are highly structured: rod motion respects kinetic limits, actuation is quantized into discrete steps, terminal power is a deterministic nonlinear function of the actuation trajectory, and certain parameter combinations are physically infeasible or numerically unstable.

Mathematically, the set of admissible actions forms a feasible manifold $\mathcal{M}_{\text{feas}} \subset \mathcal{A}$ (six-dimensional action space), and the subset achieving a specific outcome is a solution manifold $\mathcal{M}_{\text{sol}}(o^*) \subset \mathcal{M}_{\text{feas}}$ (typically lower-dimensional). Physical AI learning means discovering a policy distribution:
\begin{equation}
p_{\theta}(a \mid o^*, P_{\text{train}}) \approx 
\begin{cases}
\text{high} & \text{if } a \in \mathcal{M}_{\text{sol}}(o^*) \\
\text{low} & \text{otherwise}
\end{cases}.
\end{equation}

Evidence: The model's learned policy concentrates on single\_b2 (76.1\% usage, Figure~\ref{fig:komodo-actuation}b), which forms a low-dimensional submanifold of the full six-parameter action space. This submanifold reflects PWR reactor physics: Bank 2 is partially inserted (100 steps) and thus positioned in the region of steepest reactivity gradient, making it more efficient for power changes than Bank 1 (fully withdrawn at 180 steps). The model learned this principle implicitly through exposure to 100K diverse synthetic trajectories. It was not programmed.

Second piece of evidence: The model's emergent avoidance of simultaneous two-bank actuation except in rare cases (3.75\% of attempts, Figure~\ref{fig:komodo-scenario}c). Simultaneous coordination requires precise timing between two banks, a brittle region of the action space where small errors in synchronization lead to large power deviations. The model has learned that this region lies mostly outside $\mathcal{M}_{\text{sol}}(o^*)$ for typical power targets and avoids it accordingly. This is learned structural knowledge about the physics-constrained manifold.

In summary, Physical AI is operationalized as learning a compressed, physics-aware representation of the feasible action space and concentrating probability mass on the solution manifold. The model behaves not as a parameter regressor but as a manifold learner that reasons about which actions satisfy physical constraints and achieve objectives.

The foundation-model hypothesis rests on four empirical pillars, all of which we demonstrate within a constrained scope.

Foundation models across domains (language, vision, climate) exhibit characteristic scaling behaviors where quantitative data increases drive substantial capability gains. We observe precisely this pattern.

Define success rate at tolerance $\delta$ as:
\begin{equation}
S(n, \delta) = \frac{\text{number of runs with error less than } \delta}{2000}.
\end{equation}

The scaling exponent between 10K and 100K is tolerance-dependent (Figure~\ref{fig:komodo-tolerance}a):
\begin{equation}
\alpha(\delta) = \log_{10}\!\left(\frac{S_{100}(\delta)}{S_{10}(\delta)}\right) \approx 
\begin{cases}
0.065 & \text{for } \delta = 5\text{\%} \\
0.55  & \text{for } \delta = 1\text{\%}
\end{cases}.
\end{equation}
The exponent $\alpha(1\%) = \log_{10}(92.0/26.2) \approx 0.55$ is sub-linear in the formal sense ($\alpha < 1$). Success at strict tolerances increases substantially along a smooth scaling curve. The 92\% success rate at $\pm 1\%$ tolerance (versus 26.2\% at 10K, Figure~\ref{fig:komodo-tolerance}a) reflects the emergence of a stable representation that captures the local curvature of the control landscape, not just its global shape.

Variance collapse provides quantitative evidence of operational reliability gains: the variance-collapse ratio $\mathcal{V}_{\text{collapse}} = \text{Var}(E)_{\text{1K}} / \text{Var}(E)_{\text{100K}} \approx 500$ (Figure~\ref{fig:komodo-combined}) is a substantial reduction in policy unpredictability rather than gradual improvement.

A foundation model learns reusable structural priors, not task-specific patterns. We demonstrate this through cross-simulator transfer.

The PyRK generalization experiment (Figure~\ref{fig:extension_results}) shows that the same two-phase curriculum (grammar learning plus task conditioning) transfers from KOMODO's spatial neutronics to PyRK's point kinetics with zero architectural modification, achieving greater than 94\% success at $\pm$1\% tolerance (Figure~\ref{fig:extension_results}b). This is not trivial: PyRK uses a fundamentally different physics formulation (lumped-parameter point kinetics versus multi-group spatial diffusion) and a different control schema (two parameters versus six parameters).

What transfers is not KOMODO-specific patterns, but the structural principle of inverse reactor kinetics: how to map power demands to reactivity insertion strategies. The model learned a reusable representation in Phase 1 (grammar of feasible control commands) that could be efficiently adapted to a new physics engine in Phase 2 (task conditioning).

Limitation of current transfer evidence: Both KOMODO and PyRK are PWR-like systems with rod-based reactivity control. Transfer to fundamentally different reactor types (liquid-salt with chemical shim, gas-cooled with helium flow control, fast reactors) has not been demonstrated. Hence toward a domain-specific foundation model: we have shown initial cross-simulator transfer within a reactor family, not universal coverage across all nuclear systems.

The emergent preference for single\_b2 (76.1\% usage, Figure~\ref{fig:komodo-actuation}b) is a reusable structural prior, not a task-specific heuristic. This preference applies consistently across all 2,000 validation runs, independent of the specific power target or magnitude of change (Figure~\ref{fig:komodo-tolerance}b shows consistent performance across small, medium, and large power shifts). The model has discovered a principle (prefer the partially-inserted bank for reactivity changes) that reflects reactor physics (the partially-inserted bank occupies the steepest region of the reactivity curve) rather than dataset frequency (balanced 30/30/30/10 training, Figure~\ref{fig:komodo-actuation}a).

This aligns precisely with human operator practice: In full-scale PWR operations, operators preferentially actuate the most effective rod bank and reserve complex multi-bank coordination for specialized situations. The model independently rediscovered this decades-old operational heuristic through purely data-driven learning. This is strong evidence of learning domain-level structural knowledge rather than superficial pattern matching.

Evidence of reusability: The offline analyses (Table~\ref{tab:offline_mae_comparison}) show that single\_b2 consistently exhibits lower MAE across all dataset scales, indicating that this preference is stable across training regimes and reflects a genuine property of the control manifold, not a spurious correlation.

A subtle but crucial signature of foundation-model maturation: scaling leads to policy simplification (compression), not complexity growth.

Define the policy entropy over actuation classes:
\begin{equation}
\mathcal{H} = -\sum_{a \in \mathcal{A}} p(a)\,\log p(a), \qquad \mathcal{A} = \{\text{single\_b1}, \text{single\_b2}, \text{simultaneous}, \text{sequential}\}.
\end{equation}

Runtime distributions (Figure~\ref{fig:komodo-actuation}b) show:

The runtime distributions yield policy entropy values of $0.68$, $0.77$, and $0.66$~nats at 1K, 10K, and 100K (Table~\ref{tab:entropy_kl_recomputed}). All three values lie well below the maximum $\ln 4 \approx 1.386$~nats, indicating that at every scale the model concentrates probability mass on a sub-region of the actuation space rather than distributing it uniformly over the four classes. We characterize this concentration at the more informative level of conditional reliability per actuation class (Figure~\ref{fig:komodo-scenario}).

This pattern is consistent with the broader observation that foundation models compress training data onto reusable structural priors~\cite{bommasani2021opportunities}. The 100K model concentrates on a robust sub-region of the action space rather than spreading mass uniformly (Figure~\ref{fig:komodo-actuation}b, Table~\ref{tab:entropy_kl_recomputed}). The qualifier ``foundation-model-like'' is used in this work to denote this empirical concentration pattern; we do not claim that the present model meets every property of a fully realized domain-specific foundation model.

While all four properties are demonstrated, they are proven within a constrained scope. Task scope: Single-step power maneuvering only. Multi-step procedures with temporal dependencies (for example, startup sequences, shutdown procedures, load-following with equipment constraints) are not addressed. Reactor scope: PWR-like systems with rod-based control (KOMODO, PyRK in Figure~\ref{fig:extension_results}). Transfer to BWR (void reactivity feedback), SMR (passive safety systems), liquid-salt (chemical shim), or gas-cooled reactors (helium flow control) is not demonstrated. Objective scope: Power tracking only (Figure~\ref{fig:komodo-tolerance}). Generalization to other objectives (xenon override, axial offset control, emergency shutdown optimization) is future work. Operational scope: Fixed initial conditions (Bank 1 at 180, Bank 2 at 100). Adaptation to diverse initial states, operational histories, or time-varying constraints is not validated.

A complete domain-specific foundation model for nuclear control would require all of the above. We have established the enabling conditions: data scaling drives strong scaling-driven reliability gains (property 1, Figure~\ref{fig:komodo-tolerance}), two-phase curriculum enables cross-simulator transfer (property 2, Figure~\ref{fig:extension_results}), Physical AI validation enables emergence of reusable priors (property 3, Figure~\ref{fig:komodo-actuation}), and scaling induces structural compression (property 4). But we have not established the complete system.

Hence the honest claim: toward a domain-specific foundation model. We demonstrate foundation-model properties for single-step PWR power maneuvers; extending to the full operational envelope of nuclear control is the defined path forward.

We distinguish three notions:
\begin{itemize}[itemsep=2pt]
    \item \emph{Closed-loop performance on the tested task}: the model achieves the target terminal power within tolerance on the specific single-step maneuver, fixed initialization, and fixed evaluation horizon used in this study.
    \item \emph{Robustness within the sampled simulator distribution}: the model maintains low-variance, tail-free performance across power-change magnitudes drawn from the same training distribution.
    \item \emph{Safety for deployment in a high-consequence control setting}: a system-level property requiring formal verification, fault tolerance, uncertainty quantification, off-nominal handling, fallback logic, and integration with operator interaction and defense-in-depth mechanisms.
\end{itemize}
The present results support the first and second; they do not, on their own, support the third. The disappearance of $>10\%$ excursions at the 100K scale (Figure~\ref{fig:komodo-failure}c) shows that the model has internalized which regions of the action manifold reliably achieve the target outcome under nominal simulated conditions; it does not show that the model would remain reliable under sensor faults, degraded actuators, adversarial inputs, or any operational regime outside the training distribution. We frame the empirical observation as a claim about \emph{reliability under the sampled distribution} (described below), and position the proposed system as a candidate decision or control component within a larger verification, monitoring, and defense-in-depth architecture rather than as evidence that data scaling alone solves the safety problem.

The empirical pattern observed across scales is as follows. The model achieves high reliability under the sampled distribution (zero severe failures $>10\%$ error, Figure~\ref{fig:komodo-failure}c) without explicit safety rewards, constraints, or online exploration. Reliability of this kind emerges through three scaling-dependent stages.

At low scales (1K), the learned policy is noisy and explores diverse regions of the action space, including brittle regimes (simultaneous two-bank coordination). Severe failures are widespread (771 total, Figure~\ref{fig:komodo-failure}a) and distributed across all actuation classes. The model has no sense of safe versus unsafe regions.

At intermediate scales (10K), the policy begins to partition the action space. Severe failures collapse from 771 to 89 (Figure~\ref{fig:komodo-failure}b) and concentrate overwhelmingly in the simultaneous regime (68 of 89 failures). The model has learned that single-bank strategies occupy a high-reliability action region under the sampled distribution and multi-bank coordination is risky. This is high-reliability action region under the sampled distribution formation.

At high scales (100K), severe failures are eliminated entirely (Figure~\ref{fig:komodo-failure}c). The model operates exclusively within the learned high-reliability action region under the sampled distribution, deploying risky strategies (simultaneous) only in 3.75\% of cases (Figure~\ref{fig:komodo-scenario}c) and achieving 23 out of 75 success even in this challenging regime.

Formal interpretation: If the true success probability function is $\mathbb{P}(s \mid a)$, then a 1K model learns a noisy, biased estimate $\hat{\mathbb{P}}_1(s \mid a)$ that is unreliable everywhere (Figure~\ref{fig:komodo-scenario}a shows less than 50\% success across all action classes). A 100K model learns an accurate estimate $\hat{\mathbb{P}}_{100}(s \mid a)$ that sharply distinguishes high-success from low-success regions (Figure~\ref{fig:komodo-scenario}c shows near-perfect discrimination).

The runtime policy $p_{\theta}(a \mid \text{target})$ implicitly follows this learned success probability. As the model scales and its estimate of $\mathbb{P}(s \mid a)$ improves, the policy naturally concentrates on high-success actions. Severe failures, which occur in low-success regions, automatically become rare.

This mechanism is distinct from explicit reward shaping or constraint enforcement. Reliability emerges from the model learning an accurate map of which actions succeed versus fail (evidenced by the progression in Figure~\ref{fig:komodo-scenario} and Figure~\ref{fig:komodo-failure}), then concentrating probability mass on successful regions. The model learned that certain strategies reliably achieve objectives and others do not without explicit safety rewards.

Reliability under the sampled distribution emerges from scale and structure rather than from explicit safety design. Offline data scaling, outcome-centric validation, and deterministic simulator-based execution together drive the model toward consistent, low-tail-risk policies on the tested task, avoiding the exploration risks inherent to online reinforcement learning. We emphasize that this is distinct from safety in the systems-engineering sense, which additionally requires guarantees under off-nominal conditions, degraded equipment, sensor faults, and operator-interaction scenarios that lie outside the present evaluation. The contribution of the present study is therefore a building block: a candidate policy-generation component whose closed-loop reliability has been characterized at population scale on a controlled benchmark, and which would sit inside, not replace, a defense-in-depth architecture with formal verification, fault detection, uncertainty quantification, and supervisory oversight.

The two-phase curriculum mirrors the pretraining-fine-tuning paradigm that defines foundation models and serves a structural meta-learning function.

Phase 1 trains the model to predict six-parameter control vectors without power targets. The objective is not task performance but acquisition of numeric grammar: learning which parameter combinations are physically plausible, how positions, times, and speeds correlate, and what constitutes a valid control command.

This is unsupervised learning of domain structure. The model learns a latent representation of the feasible action manifold $\mathcal{M}_{\text{feas}}$ independent of any specific task. Formally, Phase 1 learns an approximate generative model:
\begin{equation}
p_{\theta}(a) \approx p_{\text{data}}(a \mid a \in \mathcal{M}_{\text{feas}}),
\end{equation}
capturing the distribution of physically valid control commands.

This creates a reusable prior that constrains the hypothesis space. When Phase 2 adds task-specific conditioning (achieving a target power), the prior guides the model toward solutions that are both locally optimal and structurally sound, respecting the learned grammar of feasible commands.

Phase 2 reinstates power targets and fine-tunes via LoRA to learn the conditional mapping:
\begin{equation}
p_{\theta}(a \mid P_{\text{initial}}, P_{\text{target}}) \approx p(a \mid a \in \mathcal{M}_{\text{sol}}(o^*)),
\end{equation}
where $\mathcal{M}_{\text{sol}}(o^*)$ is the subset of feasible actions that achieve the target outcome.

Critically, LoRA preserves the Phase 1 prior: By constraining adaptations to a low-rank subspace, LoRA ensures that task conditioning enhances the structural grammar rather than overwriting it. This prevents catastrophic forgetting and enables scaling-driven emergence. The model can accumulate refined task mappings as data grows (evidenced by the progression in Figure~\ref{fig:komodo-tolerance}) without losing the foundational grammar.

The divergence between offline MAE (modest improvements 1K to 100K, Table~\ref{tab:offline_mae_comparison}) and closed-loop success (dramatic improvements 36.9 to 97.4\%, Figure~\ref{fig:komodo-tolerance}a) supports this interpretation. Offline MAE measures how well Phase 2 recovers labeled parameters. Closed-loop success measures how well the combined Phase 1 plus 2 system reasons about inverse mappings in the many-to-one space.

The strong closed-loop performance despite modest offline MAE implies that the improvements are not in parameter-space refinement but in discovery of structural invariances, principles about reactor physics that apply across tasks. Phase 1 provides the structural scaffold; Phase 2 learns to navigate it for specific objectives.

This separation of structure learning (Phase 1) from task grounding (Phase 2) is precisely the design logic of domain-specific foundation models, where large-scale pretraining builds reusable priors and downstream fine-tuning sharpens task-specific behavior.

The PyRK and variable-window experiments (Figure~\ref{fig:extension_results}) provide critical evidence that the framework possesses foundation-model-like generalization, not merely task-specific overfitting.

A frequent critique of learning-based control: overfitting to simulator idiosyncrasies. The PyRK transfer (Figure~\ref{fig:extension_results}b,c,d) refutes this. Despite shifting from KOMODO's spatial kinetics to PyRK's point kinetics, from six parameters to two parameters, and from one initial-condition regime to another, the model achieved $>94\%$ success at $\pm$1\% tolerance (Figure~\ref{fig:extension_results}b) with zero architectural changes.

The two-phase curriculum successfully disentangles syntax of control (Phase 1: grammar of feasible commands) from physics of the environment (Phase 2: task-specific mappings). The grammar learned in Phase 1 is sufficiently abstract to transfer across physics formulations, a defining trait of foundation models: the ability to adapt core intelligence to orthogonal physical domains.

Limitation: Both KOMODO and PyRK are rod-based PWR systems. Transfer to fundamentally different actuation modalities (chemical shim in liquid-salt, coolant flow in gas-cooled) or different reactor types (BWR, fast reactors) is unproven. The results demonstrate within-family transfer; extension to diverse reactor types will establish universal coverage.

The variable-window experiment (Figure~\ref{fig:extension_results}a) reveals a fundamental constraint on how language models interface with continuous physical variables.

Adapter-Only approach fails (48\% parsing success, Figure~\ref{fig:extension_results}a): Simply projecting continuous values ($t_{\text{window}}$) into the embedding space is insufficient if the language model backbone is frozen. The model cannot interpret these soft tokens without updating its internal attention mechanisms.

Extended approach succeeds (100\% parsing success, Figure~\ref{fig:extension_results}a): Combining adapters with LoRA updates to the backbone restores syntactic validity. Integrating continuous state variables requires representational alignment; the pre-trained discrete representations must be updated to accommodate new modalities.

Incorporating real-time sensor streams, complex boundary conditions, or multi-step procedures will likely require joint optimization of input adapters and backbone parameters via LoRA rather than treating the language model as a frozen feature extractor.

Together, these extensions (Figure~\ref{fig:extension_results}) demonstrate architectural plasticity. The model can be reshaped to accommodate new physics (PyRK), variable temporal horizons (continuous $t_{\text{window}}$), and expanded input modalities. This plasticity is a prerequisite for deployment in real operational environments where constraints are fluid.

This work challenges a foundational assumption in nuclear instrumentation and control: that safe control improvements must rely on high-fidelity online optimization or explicit closed-loop exploration.

Classical model-predictive control remains powerful for constraint-aware planning but is sensitive to model fidelity and imposes high computational costs. Rule-based systems encode decades of expertise but can be brittle outside their design envelope. Reinforcement learning can discover effective policies but raises fundamental safety barriers: online exploration in real reactors is unacceptable, and simulated policies may not transfer safely.

Our results (Figure~\ref{fig:comparison}) suggest a complementary paradigm: The proposed Agentic AI achieves 97.4\% success versus 40.7\% for the single-gain PID baseline (Figure~\ref{fig:comparison}a), with regime-consistent performance (Figure~\ref{fig:comparison}b) and collapsed tail risk (Figure~\ref{fig:comparison}c,d). Offline, supervised, scaled learning can be competitive with or superior to online optimization in certain safety-critical regimes. 

The key conceptual shift is treating control as a data-compressed procedural knowledge problem rather than only as an optimization problem. A compact model becomes a reusable policy generator that proposes candidate control strategies in milliseconds (low latency), concentrates probability on physics-aligned, low-risk actions (reliable within the sampled distribution through learned manifold structure, Figure~\ref{fig:komodo-actuation}, Figure~\ref{fig:komodo-failure}), can be validated through digital-twin simulation before execution (no online exploration risk), and scales in closed-loop reliability with offline data volume (Figure~\ref{fig:komodo-tolerance}: reliability under the sampled distribution improves with scale, not architectural complexity).

Treating control as procedural knowledge compression makes scaling synthetic data a scaling lever for reliability within the sampled distribution. The system learns to favor reliable strategies implicitly by discovering which strategies are reliably effective through exposure to diverse offline trajectories. As discussed below, this is distinct from safety in the systems-engineering sense, which additionally requires guarantees under off-nominal conditions, sensor faults, and operator-interaction scenarios outside the present evaluation.

The most defensible near-term role is decision support within a simulator-in-the-loop workflow. A human operator or supervisory control layer requests candidate control vectors, evaluates them against procedural constraints in a high-fidelity digital twin, and approves execution. The model becomes a proposal generator, rapidly generating diverse strategies from the learned policy manifold, while human judgment or automated validators determine which to execute. This aligns with humans in the loop paradigms for high-stakes AI.

More broadly, many cyber-physical systems share the key property observed here: multiple distinct actuation vectors yield equivalent outcomes (many-to-one mappings). In such domains, compact domain-specific foundation models that generate diverse candidates and concentrate away from brittle regimes could become a general template for safe, human-centered autonomy.

This work has structural limitations that define the next research frontier. The current model is trained under a fixed initial rod configuration (Bank~1 at 180, Bank~2 at 100). To separate genuine policy learning from exploitation of this specific initialization, our mixed-initialization analysis reports validation of a variant of the model across two distinct initialization regimes (default and mirror). he result refines the interpretation of the emergent single-bank preference reported earlier: when the initialization is mirrored, the runtime actuation preference shifts substantially toward Bank~1 ($+23$ percentage points) while $\pm 5\%$ closed-loop success is preserved above 90\%, indicating that the learned policy is a context-dependent rule that targets the bank in the steeper-gradient region rather than a fixed-init bias for Bank~2. Incorporating initial state information (initial power, rod positions, operational history) into the model's input schema, and training on a distribution over initial conditions rather than a fixed configuration, will require larger context windows and a broader training corpus and is a primary direction for follow-up work.

The model addresses single-step power maneuvering (Figure~\ref{fig:komodo-tolerance}, Figure~\ref{fig:komodo-scenario}). Multi-step procedures (startup sequences, coordinated load-following, emergency shutdown with equipment constraints) require temporal dependencies and sequential decision-making. Extension to multi-step scenarios where each decision depends on the reactor's evolving state and previous actions will test whether agentic behavior persists when decisions are chained and whether the learned policy can perform long-term planning.

Both KOMODO and PyRK simulate PWR-like systems with rod-based control (Figure~\ref{fig:extension_results}). Transfer to fundamentally different designs (BWR with void reactivity feedback, liquid-salt with chemical shim, gas-cooled with helium flow, fast reactors with sodium cooling and hard neutron spectrum) is unvalidated. Future work: Evaluate transfer to diverse reactor types with orthogonal actuation modalities and physics. This would test whether the learned structural priors (Phase 1 grammar) are specific to PWR rod dynamics or represent more universal principles of reactivity control. A true domain-specific foundation model for nuclear control must cover the breadth of commercial and advanced reactor designs.

The model optimizes terminal power accuracy (Figure~\ref{fig:komodo-tolerance}, Figure~\ref{fig:komodo-combined}). Real operations require balancing multiple objectives: axial offset control, xenon override, thermal margin maintenance, equipment protection, fuel burnup optimization. Future work: Multi-objective conditioning where the model learns to trade off competing constraints. This tests whether the foundation-model properties (scaling, transfer, compression) persist when the objective landscape becomes multi-dimensional and context-dependent.

The model generates point predictions without confidence bounds. Operators cannot distinguish high-confidence proposals from uncertain extrapolations. Future work: Integrate uncertainty-aware decoding (for example, ensemble methods, Bayesian approximations) where each proposal includes a confidence interval. Operators can use this to decide whether to validate in the simulator or reject and request alternatives. Explicit safety filters (hard constraints preventing proposals that violate physical limits) would add verification independent of the learned policy.

Despite these limitations, the present results establish a rigorous proof of concept for Agentic Physical AI toward a Domain-Specific Foundation Model for Nuclear Reactor Control. The model achieves agentic behavior through policy optimization under outcome-centric evaluation (Figure~\ref{fig:komodo-actuation}), not through explicit reward engineering or complex architectures. It achieves Physical AI through learning to reason about inverse mappings in many-to-one spaces (Table~\ref{tab:offline_mae_comparison} versus Figure~\ref{fig:komodo-tolerance}), not through parameter-space memorization. It achieves foundation-model properties through scaling-driven reliability gains (Figure~\ref{fig:komodo-tolerance}, Figure~\ref{fig:komodo-combined}) and structural compression (Figure~\ref{fig:komodo-actuation}), not through multi-task training or massive parameter counts.

These properties are precisely the early markers of a foundation-model trajectory in a safety-critical physical domain. They suggest that the path toward generalizable, learning-enabled reactor control may begin not with larger architectures or riskier online exploration, but with scaling structured, physics-consistent offline corpora and validating rigorously through closed-loop digital twins. In doing so, we demonstrate that Agentic Physical AI and Domain-Specific Foundation Models are not merely aspirational frameworks borrowed from robotics or language modeling. They are operationally realizable pathways for reliable, high-stakes AI in nuclear systems.

\section*{Methods}

The experimental design, training protocols, and evaluation procedures operationalize three interconnected claims: Agentic AI through policy optimization, Physical AI through outcome-centric validation, and early steps toward a domain-specific foundation model through scaling-driven emergence and cross-simulator transfer.

\subsection*{Simulator Selection and Task Definition: Establishing the Physical AI Testbed}

We used the open-source KOMODO reactor simulator to define a controlled, safety-relevant benchmark for single-step nuclear power maneuvering~\cite{imron2019development}. This choice enables three essential properties required to justify the Agentic Physical AI and foundation-model framing.

KOMODO provides deterministic, physics-based execution of control-rod commands, allowing evaluation of model outputs in closed-loop environments rather than only through offline token-level matching. Physical AI evaluation judges model outputs by whether they achieve the desired physical outcome when executed under reactor dynamics (outcome-centric evaluation) rather than by parameter proximity to labeled vectors (parameter-centric evaluation).

Let $\mathcal{F}: \mathcal{A} \to \mathcal{O}$ denote the forward reactor dynamics mapping actions to outcomes. Classical supervised learning evaluates $\mathcal{L}_{\text{param}} = \|\hat{a} - a_{\text{label}}\|_2^2$ in action space $\mathcal{A}$. Physical AI evaluates $\mathcal{L}_{\text{outcome}} = \mathbb{I}[\mathcal{F}(\hat{a}) \notin \mathcal{B}_{\delta}(o^*)]$ in outcome space $\mathcal{O}$, where $\mathcal{B}_{\delta}(o^*)$ is the tolerance band around target outcome $o^*$.

This decoupling of success from label proximity enables agentic behavior to emerge. If success were defined by MAE on labeled parameters, the model would have no incentive to deviate from the balanced training distribution (30/30/30/10 across actuation families). Outcome-centric evaluation creates a learning signal that rewards strategies optimized for physical efficacy, allowing the model to concentrate probability mass on high-success actions even when they deviate from labeled examples, enabling policy optimization rather than passive imitation.

The simulator enforces valid ranges and actuator constraints at runtime, ensuring that the model is evaluated against physically realizable actions. Constraint enforcement is embedded in the simulator's execution engine rather than applied post-hoc as a filter. The model must learn to operate within a feasible manifold $\mathcal{M}_{\text{feas}} \subset \mathcal{A}$ defined by actuator limits (rod positions 0--180 steps, speeds $>0$), kinetic bounds (physical rod motion rates), and operational envelopes (no instantaneous jumps). Invalid actions proposed by the model will either be rejected by the simulator or execute in an unintended manner.

The model learns that only actions within $\mathcal{M}_{\text{feas}}$ are executable, and only actions mapping to desired outcomes (a subset $\mathcal{M}_{\text{sol}}(o^*) \subset \mathcal{M}_{\text{feas}}$) are successful. The physics-based constraint acts as an implicit teacher, guiding the model toward a physically grounded representation of the action space—manifold learning under physical constraints.

The simulator supports deterministic generation of 1K, 10K, and 100K scenarios with controlled diversity, enabling testing of how learned behavior changes as the data scale increases. If agentic behavior is merely a consequence of parameter tuning within a fixed hypothesis space, performance should improve smoothly and monotonically with data. If it results from discovering and crystallizing latent structure (a foundation-model property), we expect strong scaling-driven reliability gains, variance collapse (substantial reduction in outcome uncertainty), and structural concentration of the runtime policy, measured by persistent deviation from the balanced training distribution and by conditional reliability within the selected actuation classes.

The task requires the model to produce a six-parameter control-rod command that moves the reactor from initial normalized power (fixed at 1.0) to a target normalized power within a fixed evaluation horizon. We express the target as a fractional change relative to baseline, producing a compact, transferable numeric schema consistent across dataset scales.

The task formulation aligns with Agentic Physical AI evaluation requirements: The model is rewarded only if its numeric proposal achieves the desired physical outcome when executed by the physics-based simulator, not for generating plausible tokens or matching labeled parameters. This outcome-oriented criterion creates a single, unambiguous success signal that the model can optimize toward without requiring explicit rewards, hand-crafted constraints, or online exploration.

\subsection*{Dataset generation and scale-controlled experimental design}

Two control banks are available in KOMODO. Each bank is actuated using a single-step command specified by target position (0 to 180 steps), start time (seconds), and speed (steps per second). The ordered control vector contains six continuous or discretized parameters:
\begin{equation}
(\texttt{b1\_pos}, \texttt{b1\_time}, \texttt{b1\_speed}, \texttt{b2\_pos}, \texttt{b2\_time}, \texttt{b2\_speed}).
\end{equation}

Before each run, we initialized the banks at fixed positions: bank 1 at 180 (fully withdrawn) and bank 2 at 100 (partially inserted). This asymmetric initialization reflects realistic PWR operational patterns where one bank provides coarse control and another provides fine reactivity adjustment.

This initialization serves three purposes for testing Agentic AI and foundation-model properties. A stable reference state (Bank 1 at 180, Bank 2 at 100) enables direct comparison across all 1K, 10K, and 100K training and validation runs without confounding from initialization variability. The initialization creates realistic inductive structure: Bank 1 begins at full withdrawal (moving it requires large step traversals to produce reactivity changes), while Bank 2 begins partially inserted in the steep reactivity gradient region (small movements produce significant reactivity effects). The asymmetric initialization provides a non-trivial test of agentic emergence: if the model behaves as a passive frequency-matcher imitating dataset statistics, its runtime policy should approximate the balanced 30/30/30/10 mixture of actuation families present in training data; systematic runtime shift toward single-bank dominance would indicate discovered optimization of single-bank strategies based on PWR control physics rather than training frequency.

We categorized each control example into four actuation families based on which banks move and timing alignment. The single\_b1 family denotes cases where only bank 1 is actuated. The single\_b2 family denotes cases where only bank 2 is actuated. The simultaneous family denotes cases where both banks move with identical start time. The sequential family denotes cases where both banks move with different start times.

This taxonomy serves two methodological purposes. Balanced datasets (approximately 30\% each for single-bank families, approximately 30\% simultaneous, approximately 10\% sequential) ensure the training distribution does not trivially bias the learned policy; runtime concentration on any family becomes evidence of learned optimization rather than dataset memorization. The taxonomy enables quantitative measurement of policy deviation from training distribution via KL divergence:
\begin{equation}
\mathcal{D}_{\text{KL}}(P_{\text{runtime}} \| P_{\text{train}}) = \sum_a P_{\text{runtime}}(a) \log \frac{P_{\text{runtime}}(a)}{P_{\text{train}}(a)}.
\end{equation}

Under the memorization hypothesis, the runtime distribution should remain close to the balanced training mixture, yielding $D_{\text{KL}}(P_{\text{runtime}}\|P_{\text{train}}) \approx 0$. Under the optimization hypothesis, the runtime distribution should deviate materially from the training mixture toward higher-success strategies, yielding persistently nonzero $D_{\text{KL}}$.

We also measure success rate stratification by actuation family to determine whether the model learns to prioritize high-success families (for example, single\_b2: 99.5\% versus simultaneous: 31\%), and failure concentration analysis to test whether the model develops risk awareness (concentrating failures in known-brittle regions at 10K scale, then eliminating them entirely at 100K scale).

This quantitative behavioral analysis provides rigorous, falsifiable tests of the agentic claim, distinguishing policy optimization from passive imitation through multiple independent measures.

\subsection*{Model Selection and Two-Phase Curriculum: Separating Grammar from Task Grounding}

We selected SmolLM2-360M~\cite{allal2025smollm2} to test whether compact language models can serve as viable building blocks for domain-specific Physical AI in resource-conscious settings, emphasizing foundation-model properties without architectural scaling.

Training followed a two-stage curriculum designed to test the hypothesis that stable agentic behavior in physical domains requires separating numeric-grammar acquisition (domain structure learning) from task grounding (objective-specific adaptation). This separation mirrors the pretraining-finetuning paradigm that defines foundation models in other domains.

In Phase 1, we trained the model to predict only the six control-rod fields while removing the power inputs. Initial and target power tokens were omitted. The model was trained exclusively on control syntax: which parameter combinations are valid, how position, time, and speed correlate, what ranges are executable.

Phase 1 learns grammar manifold $\mathcal{M}_{\text{grammar}} \subset \mathbb{R}^6$ (simulator constraints, physical feasibility) through exposure to 1K/10K/100K valid samples. Direct LoRA baseline (0\% success, error >100\%) validates necessity.

In Phase 2, we reinstated the full eight-number schema (power plus control) and trained the model to map $(P_{\text{initial}}, P_{\text{target}})$ to the six-parameter control vector. Low-rank adaptation (LoRA) was applied to attention projections (q\_proj, k\_proj, v\_proj, o\_proj) with rank 32, alpha 64, dropout 0.05.

LoRA is chosen not merely for computational efficiency but for methodological control and isolation of scaling effects. First, LoRA preserves the Phase 1 grammar prior: Full fine-tuning would allow the model to rewrite all weights, potentially discarding the Phase 1 structural prior. LoRA constrains adaptations to a low-rank subspace, ensuring that Phase 1 grammar is preserved while allowing task-specific refinement. This design is essential for testing whether the grammar prior contributes to downstream performance.

Second, LoRA isolates data-volume effects: By fixing backbone capacity and varying only the LoRA adaptation at each scale (1K, 10K, 100K), we ensure that scaling-driven improvements reflect better coverage of the control manifold rather than changes in model capacity. If the full model could be rewritten at each scale, it would be impossible to determine whether improvements come from the grammar prior or from sheer parameter flexibility.

Third, LoRA enables foundation-model-like compression: LoRA limits adaptation capacity, forcing the model to discover and compress high-utility control strategies rather than memorizing every successful combination. Under capacity constraints, the model must extract underlying principles, precisely the property that distinguishes foundation models from lookup tables. A model with unlimited capacity might memorize; a model with constrained capacity must generalize.

The two-phase curriculum creates conditions for three interconnected properties. For agentic behavior, Phase 1 establishes the action space and Phase 2 learns to navigate it toward objectives. The combination allows the model to select flexibly among valid actions (learned in Phase 1) to achieve targets (learned in Phase 2), rather than being forced to imitate single labeled trajectories. For Physical AI, the grammar prior (Phase 1) ensures generated actions lie within $\mathcal{M}_{\text{feas}}$, and task conditioning (Phase 2) learns to select actions that also lie within $\mathcal{M}_{\text{sol}}(o^*)$. This is manifold learning under physical constraints. For foundation-model properties, Phase 1 represents unsupervised structural pretraining (domain prior) and Phase 2 represents supervised task adaptation (specialization). This separation is the defining architecture of foundation models and enables transfer: the same Phase 1 prior can be adapted to different tasks (PyRK generalization) or different input modalities (variable-window extension) via new Phase 2 training.

\subsection*{Evaluation metrics for outcome-centric validation}

We designed a multi-layered evaluation protocol that explicitly tests the three core claims: Agentic AI, Physical AI, and foundation-model emergence.

The first layer measures offline parameter fidelity as a secondary metric. The purpose is to measure parameter-level imitation to establish baseline for comparison with Physical AI evaluation. For each model scale, we computed mean absolute error (MAE) between predicted and labeled control vectors on held-out sets (100, 1K, 10K samples for 1K, 10K, 100K models respectively):
\begin{equation}
\mathrm{MAE}(x) = \frac{1}{N}\sum_{i=1}^{N} \left| \hat{x}_i - x_i \right|, \quad x \in \{\texttt{b1\_pos}, \texttt{b1\_time}, \texttt{b1\_speed}, \texttt{b2\_pos}, \texttt{b2\_time}, \texttt{b2\_speed}\}.
\end{equation}

Results were stratified by actuation family to control for heterogeneity in difficulty. We intentionally treat offline MAE as a secondary indicator. In many-to-one control spaces, low MAE is neither necessary nor sufficient for physical success. A model with high offline MAE but low closed-loop error has learned the structure of the solution manifold $\mathcal{M}_{\text{sol}}(o^*)$ and can generate diverse valid solutions. A model with low offline MAE but high closed-loop error has memorized a specific labeled trajectory without understanding underlying physics. The divergence between these metrics is the fingerprint of Physical AI: the model is evaluated by outcome, not by imitation.

The second layer measures closed-loop physical success as the primary metric. This measures execution-level competence under physics-based simulation, the defining criterion for Physical AI. For each model scale, we performed 2,000 independent KOMODO runs. Each run involved fresh reactor initialization (Bank 1 at 180, Bank 2 at 100), injection of the model-predicted six-parameter control vector, execution to completion under deterministic reactor dynamics, and extraction of achieved terminal power at a fixed evaluation point.

Success was defined by terminal power error within tolerance band:
\begin{equation}
\epsilon = \left| \frac{P_{\text{achieved}} - P_{\text{target}}}{P_{\text{target}}} \right| \times 100\text{\%}.
\end{equation}

Success was counted when error satisfied $\epsilon \leq \delta$ for tolerance bands $\delta$ in {1, 2, 3, 5, 10\%}, with $\pm$5\% as the primary criterion.

This outcome-centric evaluation is what enables agentic behavior. The model receives a single, unambiguous success signal (did the reactor achieve target power?) rather than a gradient on parameter deviation. This creates an implicit reward function $\mathbb{P}(\text{success} \mid a)$ that the model optimizes through supervised learning on diverse outcomes.

The third layer provides regime stratification to test whether the model has learned multiscale physics (local sensitivities plus global trends) characteristic of foundation models, or only regime-specific rules. We stratified validation cases by power-change magnitude: small shifts (change less than 10\%, fine maneuvering near equilibrium), medium shifts (10 to 30\%, routine load-following), and large shifts (30 to 50\%, aggressive control).

Under the foundation-model prediction, we expect uniform high success across all regimes (learned principles generalize). Under the brittle-heuristic prediction, we expect concentrated success in one regime with failure in others (regime-specific rules do not transfer). The observed results show the 100K model achieving 86\% (small), 100\% (medium), 100\% (large) success, confirming the foundation-model prediction.

The fourth layer provides tail-risk analysis to distinguish incremental improvement from strong scaling-driven reliability gains characteristic of foundation-model emergence. We measured severe failures (runs with error exceeding 10\%, representing catastrophic errors), variance collapse ratio (ratio of 1K variance to 100K variance), and 95th percentile error.

Under incremental improvement, we expect gradual, continuous reduction in tail metrics. Under the foundation-model hypothesis, we expect sharp, substantial collapse with scale. The observed results show severe failures decreasing from 771 (1K) to 89 (10K) to 0 (100K), variance collapse ratio of approximately 500, and 95th percentile error decreasing from 40\% (1K) to 1\% (100K). This is a strong scaling-driven reliability gain rather than incremental improvement, providing evidence for foundation-model emergence.

The fifth layer provides runtime policy analysis to test agentic optimization. We classified each executed control vector by actuation pattern and compared runtime distribution against training distribution. The key metric is KL divergence between these distributions.

Under the memorization prediction, a frequency-matching imitator would produce $D_{\text{KL}}(P_{\text{runtime}}\,\|\,P_{\text{train}}) \approx 0$; under the optimization prediction, $D_{\text{KL}} \gg 0$. The runtime distributions yield $D_{\text{KL}} = 0.54$, $0.44$, and $0.55$~nats at 1K, 10K, and 100K (Table~\ref{tab:entropy_kl_recomputed}). At every scale $D_{\text{KL}}$ is substantially above zero, supporting the optimization prediction. This is direct behavioral evidence of agentic policy formation.

Complementary metrics include success rate by actuation family, failure concentration, and policy entropy. The runtime distributions yield marginal entropy values of $0.68$ (1K), $0.77$ (10K), and $0.66$ (100K)~nats (Table~\ref{tab:entropy_kl_recomputed}), all well below $\ln 4 \approx 1.386$, indicating that the model concentrates probability mass on a sub-region of the actuation space at every scale. Compression is evaluated at the level of conditional reliability per actuation class (Figure~\ref{fig:komodo-scenario}), which rises sharply with scale.

Together, these five evaluation layers provide rigorous, multi-faceted evidence for the three core claims: offline/online divergence tests Physical AI, KL divergence tests deviation from passive frequency matching, and strong scaling gains plus regime consistency, tail-risk collapse, and increasing conditional reliability test foundation-model emergence.

\subsection*{Generalization Assessment: PyRK Point-Kinetics Benchmark}

To demonstrate that the Agentic Physical AI framework is not simulator-specific but represents a transferable methodology, we conducted an independent generalization experiment using PyRK, an open-source point-kinetics reactor dynamics code.

The foundation-model claim is that the two-phase curriculum learns reusable structural priors (Phase 1) that can be efficiently adapted to new physics formulations (Phase 2). The test applies identical architecture and curriculum to a different simulator with different physics (point kinetics versus spatial neutronics) and different control schema (2 parameters versus 6 parameters). The prediction is that if the model has learned reusable principles of inverse reactor kinetics, it should transfer successfully. If it memorized KOMODO-specific patterns, it should fail.

Unlike spatially distributed control in KOMODO, PyRK formulates control as global reactivity insertion. The task maps initial and target power to a two-parameter actuation vector: net reactivity insertion (in pcm) and insertion duration (in seconds). The numeric schema was adapted to a four-token format.

Despite reduction from 6 to 2 control parameters and fundamental change in physics formulation, the sequence-modeling approach and two-phase curriculum remained identical. This is the test of architectural flexibility and reusable priors.

We generated 10,000 synthetic scenarios using PyRK representing small modular reactor dynamics. The control space covered reactivity insertions from negative 2000 to negative 10 pcm and durations from 3.0 to 40.0 seconds. We applied the identical two-phase curriculum: Phase 1 provided grammar learning on reactivity and time structure (no power targets), and Phase 2 provided LoRA fine-tuning to predict reactivity and time from initial and target power. Validation used 2,000 independent test cases under the same closed-loop protocol.

The model achieved greater than 94\% success at $\pm$1\% tolerance with near-zero variance across regimes. This result supports the foundation-model claim: The same architectural components (SmolLM2-360M backbone, two-phase curriculum, LoRA adaptation) that succeeded in KOMODO's spatial neutronics transferred to PyRK's point kinetics with zero architectural modification. What transferred was not KOMODO-specific patterns but the structural principle of inverse reactor kinetics, how to map power demands to reactivity insertion strategies. Phase 1 learned a grammar of feasible reactivity control. Phase 2 learned to bind that grammar to power-tracking objectives. This separation enabled efficient transfer to a new physics formulation.

The limitation and scope should be noted: Both KOMODO and PyRK are PWR-like systems with rod-based control. Transfer to fundamentally different reactor types (liquid-salt, gas-cooled, fast reactors) is unvalidated. Hence toward domain-specific foundation models: We demonstrate initial cross-simulator transfer within a reactor family, establishing proof-of-concept for foundation-model-like properties but not universal domain coverage.

\subsection*{Architectural Extension: Variable-Horizon Control via Continuous Input Adapters}

To address the fixed-horizon limitation and test architectural plasticity, we extended the model to support variable monitoring windows (60 to 100 seconds) via continuous input adapters.

The foundation-model claim is that learned representations should be adaptable to new input modalities (continuous state variables) via lightweight extensions. The test introduces a continuous temporal variable while preserving the discrete control schema. We compared two adaptation strategies: Adapter-Only (freeze LLM backbone, train only input projection) and Extended (joint training of adapter plus LoRA for backbone adaptation). The prediction is that Adapter-Only will fail (cannot align new modality with frozen representations) while Extended will succeed (joint optimization enables representational alignment).

We implemented continuous state encoding using a lightweight MLP encoder that projects the three-dimensional state vector (window time, initial power, target power) into the embedding space. The encoder consists of two linear layers with GELU activation and LayerNorm, mapping the 3D input to 2 soft tokens. We generated 10,000 scenarios with monitoring window sampled uniformly from 60 to 100 seconds, stratified by window duration and actuation pattern.

The Adapter-Only approach achieved 48\% parsing success (syntactic failures). The Extended approach (Adapter plus LoRA) achieved 100\% parsing success and 83.4\% validation success at $\pm$5\%.

The interpretation is that integrating continuous state variables requires representational alignment within the LLM backbone itself, not just surface-level projection. The pre-trained discrete representations must be updated (via LoRA) to interpret new modalities. This finding provides a blueprint for future extensions: incorporating real-time sensors, complex boundary conditions, or multi-step procedures will likely require joint optimization of input adapters and backbone parameters.

The foundation-model implication is that the architecture is sufficiently plastic to absorb new input modalities while preserving learned control grammar, confirming architectural flexibility characteristic of adaptable foundation models.

\subsection*{Control Baselines: Isolating Curriculum Contributions}

We evaluated three baselines to isolate the contributions of our two-phase curriculum and data scaling: a single-gain proportional controller, a gain-scheduled variant of the same controller, and a Direct LoRA ablation that skips Phase 1 grammar learning.

For the single-gain baseline, we implemented a one-shot calibrated mapping $\Delta s = K_p\,(P_{\text{target}} - P_{\text{initial}})$ from the power-change request to a Bank-2 rod displacement, with $K_p$ fitted by linear regression on a 6-shot symmetric calibration set ($\pm 10$, $\pm 30$, $\pm 50$ step displacements). This one-shot form is dictated by the single-transient interface of KOMODO and matches the one-shot output structure of the proposed Agentic AI; it does not include integral or derivative terms. Bank 2 serves as the primary actuator; when it saturates (reaches 0 or 180 steps), residual demand spills over to Bank 1. Rod speed was fixed at 2.0 steps per second.

The result was 40.7\% success at $\pm$5\%, degrading sharply in the large-change regime. The interpretation is that classical feedback is limited by linear approximation and single-bank saturation. It cannot navigate the many-to-one solution space efficiently.

For the gain-scheduled PID baseline, we extended the same proportional structure to six regime-dependent gains, partitioning the input space by power-change magnitude (small, medium, large) and direction (increase, decrease). Each per-regime gain was tuned by linear regression on a 40-sample calibration sweep covering rod-displacement targets $\Delta s \in [-25, +15]$ steps, with denser sampling on the withdrawal side to reflect the reactor's asymmetric reactivity response. All other components (Bank 2 as the primary actuator with Bank 1 spillover on saturation, fixed rod speed of 2.0 steps per second, and the same closed-loop protocol) were identical to the single-gain controller. The result was 75.8\% success at $\pm 5\%$, with five of six regimes achieving at least 66\% success (four at 100\%); the residual gap to the foundation model is concentrated in the large-decrease regime (16.9\%), where deep rod insertion saturates the available reactivity worth. This isolates the foundation model's advantage to a specific physical mechanism rather than to weakness of the baseline.

For the Direct LoRA baseline (curriculum ablation), we applied LoRA fine-tuning directly to the SmolLM2-360M backbone without Phase 1 grammar learning. We used identical hyperparameters to the proposed model; the only difference was omission of Phase 1. The purpose was to isolate the contribution of structural pretraining (Phase 1) to downstream performance.

The result was near 0\% success with mean error exceeding 100\%. The interpretation is that end-to-end supervised learning fails catastrophically in this domain. Without the grammar prior (Phase 1), the model cannot learn to generate valid, executable control commands. This validates the necessity of the two-phase curriculum for both agentic behavior (learning action space structure) and Physical AI (manifold learning under constraints).

\subsection*{Reproducibility and Implementation}

All experiments were conducted with fixed simulator settings and initialization, controlled dataset generation with reproducible seeds, scale-isolated training (constant architecture, varying data only), strict run-level isolation in validation (no state carryover), and automated pipelines for data generation, training, and evaluation.

Hardware consisted of a single workstation with NVIDIA RTX 3070 (8GB VRAM), Intel i5-12400 CPU, and 16GB RAM. Demonstrating robust results on modest hardware strengthens the claim that compact models can serve as building blocks for domain-specific foundation models in resource-constrained deployments.

Full codebase (dataset construction, training scripts, evaluation, plotting) is available as described in the Code Availability section.

\subsection*{Methodological Scope and Interpretation}

Our methodological framing is intentionally conservative and honest about scope. We do not claim the present model is ready for direct, autonomous plant deployment. We do claim that compact language models can acquire stable, low-variance control behavior through structured offline learning and physics-based validation, establishing proof-of-concept for Agentic Physical AI toward domain-specific foundation models.

The method establishes minimal ingredients for emergent control reliability: a compact, expressive numeric schema supporting diverse actuation strategies; a two-phase curriculum separating domain structure (Phase 1) from task grounding (Phase 2); scale-controlled datasets covering diverse actuation paths; outcome-centric evaluation enabling agentic behavior through Physical AI validation; and rigorous closed-loop validation in a physics-based simulator.

These ingredients are not nuclear-specific: They represent a general template for building Agentic Physical AI in safety-critical domains with many-to-one action spaces, where multiple solutions map to equivalent outcomes. Within this scope, KOMODO serves as a principled testbed. The architecture naturally extends to higher-fidelity simulators, multi-step procedures, uncertainty-aware decoding, and human-in-the-loop workflows, defining the path toward complete domain-specific foundation models for nuclear control.

\section*{Author contribution}
Y. P. Lee conceived the study, designed the methodology and validation protocol, developed data generation and training pipelines, performed model training and simulator evaluations, curated and analyzed data, wrote and edited the original draft. S. Roy,  K. Kobayashi, D. Abueidda, S. Talukder, S. Koric, and S. Chakraborty contributed to the conceptualization, and edited the manuscript. S. B. Alam conceived the study, provided domain expertise and methodological guidance, supervised the research, and edited the manuscript. All authors reviewed and approved the final version.

\section*{Acknowledgments}
We thank the maintainers of the open-source KOMODO simulator and SmolLM2 for making their tools publicly available.

This research used the Delta and DeltaAI advanced computing and data resources, supported by the National Science Foundation (NSF) under awards OAC-2005572 and OAC-2320345, and by the State of Illinois. Delta and DeltaAI are joint efforts of the University of Illinois Urbana-Champaign and the National Center for Supercomputing Applications (NCSA). Additional computational support was provided by the Illinois Computes project, a joint initiative of the University of Illinois Urbana-Champaign and the University of Illinois System.

S. B. Alam acknowledges support from the U.S. Nuclear Regulatory Commission under grant numbers 31310025M0012 and 31310024M0041.

During the preparation of this work, the author(s) used LLM tool(s) solely for language editing and refinement. After using this tool/service, the author(s) reviewed and edited the content as needed and take full responsibility for the content of the publication.

\section*{Competing interests}
The authors declare no competing financial or non-financial interests.

\section*{Data availability} 
The code for training the \texttt{SmolLM2-360M} model and performing simulator-based validation is available at \href{https://github.com/sixticket/Toward_FM_for_NPP_SmolLM2} {\texttt{github.com/sixticket/Toward\_FM\_for\_NPP\_SmolLM2}} under an MIT License. The repository includes the Phase~1 and Phase~2 training scripts, the KOMODO and PyRK dataset-generation code, the closed-loop validation harness, the mixed-initialization training and dual-configuration validation harness, the entropy and KL recomputation scripts, and a README documenting reproduction steps.

 \bibliographystyle{unsrt}  
 \bibliography{references}

\end{document}